\documentclass[conference]{IEEEtran}
\usepackage{tabularx}
\usepackage{array, boldline, booktabs, makecell}
\usepackage{multirow}
\usepackage{amsmath,amssymb,amsfonts}
\usepackage{algorithmic}
\usepackage{graphicx}
\usepackage{textcomp}
\usepackage{caption}
\usepackage{xcolor}
\usepackage{hyperref, etoolbox}
\usepackage{float}
\usepackage{bm}
\usepackage{amsmath,amsfonts,amsthm}
\usepackage{mathrsfs}
\def\BibTeX{{\rm B\kern-.05em{\sc i\kern-.025em b}\kern-.08em
    T\kern-.1667em\lower.7ex\hbox{E}\kern-.125emX}}
\renewcommand{\arraystretch}{1.15}
\newcolumntype{P}[1]{>{\centering\arraybackslash}p{#1}}
\newcolumntype{M}[1]{>{\centering\arraybackslash}m{#1}}

\usepackage{booktabs}
\usepackage{tabularx, ragged2e}
\usepackage[linesnumbered,ruled]{algorithm2e}
\usepackage{verbatim}
\usepackage{fancyvrb}
\usepackage{fvextra}

\usepackage{amsmath,amsfonts,bm}

\def\eqref#1{equation~(\ref{#1})}

\def\1{\bm{1}}

\def\vk{{\bm{k}}}

\def\vp{{\bm{p}}}

\def\vx{{\bm{x}}}

\DeclareMathAlphabet{\mathsfit}{\encodingdefault}{\sfdefault}{m}{sl}
\SetMathAlphabet{\mathsfit}{bold}{\encodingdefault}{\sfdefault}{bx}{n}

\def\gD{{\mathcal{D}}}

\def\gX{{\mathcal{X}}}
\def\gY{{\mathcal{Y}}}

\newcommand{\E}{\mathbb{E}}
\newcommand{\Ls}{\mathcal{L}}

\DeclareMathOperator*{\argmin}{arg\,min}

\begin{document}

\title{Backdoor Attacks on Time Series: A Generative Approach}

\author{\IEEEauthorblockN{Yujing Jiang\textsuperscript{1}, Xingjun Ma\textsuperscript{2}, Sarah Monazam Erfani\textsuperscript{1}, James Bailey\textsuperscript{1}}
\IEEEauthorblockA{\textsuperscript{1}\textit{Faculty of Engineering and Information Technology, The University of Melbourne}\\
\{yujingj@student., sarah.erfani@, baileyj@\}unimelb.edu.au}

\IEEEauthorblockA{\textsuperscript{2}\textit{School of Computer Science, Fudan University}\\
xingjunma@fudan.edu.cn}
}

\maketitle

\begin{abstract}
Backdoor attacks have emerged as one of the major security threats to deep learning models as they can easily control the model's test-time predictions by pre-injecting a backdoor trigger into the model at training time. While backdoor attacks have been extensively studied on images, few works have investigated the threat of backdoor attacks on time series data. To fill this gap, in this paper we present a novel generative approach for time series backdoor attacks against deep learning based time series classifiers. Backdoor attacks have two main goals: high stealthiness and high attack success rate. We find that, compared to images, it can be more challenging to achieve the two goals on time series. This is because time series have fewer input dimensions and lower degrees of freedom, making it hard to achieve a high attack success rate without compromising stealthiness. Our generative approach addresses this challenge by generating trigger patterns that are as realistic as real-time series patterns while achieving a high attack success rate without causing a significant drop in clean accuracy. \footnote{\href{https://github.com/yujingmarkjiang/Time\_Series\_Backdoor\_Attack}{https://github.com/yujingmarkjiang/Time\_Series\_Backdoor\_Attack}.} We also show that our proposed attack is resistant to potential backdoor defenses. Furthermore, we propose a novel universal generator that can poison any type of time series with a single generator that allows universal attacks without the need to fine-tune the generative model for new time series datasets.
\end{abstract}

\section{Introduction}

Time series captures a sequence of observations with measurable quantities indexed by timestamps. It is amongst the most ubiquitous data types in a wide range of industries, such as finance \cite{peia2015finance}, heavy industry \cite{essien2020deep}, and healthcare \cite{penfold2013use, kaushik2020ai}. Similar to the computer vision field, deep neural networks (DNNs) are often used in time series analysis to achieve state-of-the-art performance \cite{gamboa2017deep,wang2017time,zhao2017convolutional}. However, DNNs are known to be vulnerable to backdoor attacks where the adversary aims to control the model's test-time prediction behaviors by implanting a backdoor trigger into the model at training time \cite{gu2017badnets, liu2020reflection}. This has raised security concerns with the deployment of DNN models in safety-critical applications.

Backdoor attacks represent one type of training-time vulnerabilities of DNNs, which are different from test-time adversarial attacks~\cite{szegedy2013intriguing,goodfellow2014explaining,ma2018characterizing}. To implant the backdoor trigger into a target model, the adversary can either poison a small fraction of the training data with a trigger pattern \cite{gu2017badnets} or directly manipulate the training procedure \cite{liu2020reflection}. The former could occur during the data collection process, while the latter could happen to outsourced model training or the use of pre-trained models downloaded from untrusted sources. A backdoored model predicts the correct classes on clean test inputs yet will constantly predict the backdoor class whenever the trigger pattern appears.

Backdoor attacks have been extensively studied on images with DNN-based image classifiers, however, few works have investigated the potential backdoor vulnerability of DNN-based time series models. 
The two main objectives of backdoor attacks, namely high attack success rate (ASR) and high stealthiness, have been achieved on images as demonstrated by many existing works \cite{gu2017badnets,chen2017targeted,turner2019label,liu2020reflection,cheng2020deep}. 
The effectiveness of backdoor attacks is closely associated with their trigger patterns, which are often designed to be fixed patterns or dynamic but sparsely distributed patterns for images.
However, unlike images, time series are generally of lower dimensions (e.g., univariate) and fewer degrees of freedom (e.g., limited window length). It thus makes fixed patterns more noticeable (less stealthy) on time series.
In fact, it is still unclear whether fixed patterns are effective on time series. Moreover, time series are of diverse types, such as stock prices, temperature readings, weather data, and heart rate monitoring, to name a few. As such, fixed patterns can hardly be effective on all types of time series.

In this paper, we present a novel generative approach for generating stealthy and sample-specific trigger patterns for effective time series backdoor attacks. By leveraging generative adversarial networks (GANs), our approach can generate backdoored time series (the original time series plus the trigger pattern) that are as realistic as real-time series, while achieving a high attack success rate. Furthermore, by training the trigger pattern generator on multiple types of time series, we can obtain a universal generator.
The universal trigger generator demonstrates high flexibility in performing backdoor attacks on different types of time series across different domains, revealing the significant threat of backdoor attacks to time series analysis. Our work provides a practical solution to stealthy and effective time series backdoor attacks, and reveals the potential backdoor vulnerability of DNN-based time series classification models.

In summary, our main contributions are:

\begin{itemize}
\item We study the problem of backdoor attacks on time series and propose a novel generative approach for crafting stealthy sample-specific backdoor trigger patterns. We also reveal the unique challenge of time series backdoor attacks posed by the inherent properties of time series, (\textit{i.e.,} low dimension and limited degrees of freedom).
\item We empirically show that our proposed attack can generate stealthy and effective backdoor attacks against state-of-the-art DNN-based time series models and is resistant to potential backdoor defenses. The attacked models also have minimal clean accuracy drop on both univariate and multivariate datasets.
\item We present a novel universal backdoor attack that is capable of crafting sample-specific backdoor triggers for different types of time series across a wide range of domains. With a one-time training on a combination of time series datasets, the proposed universal attack can succeed 70\% of the time under the poison-label setting.
\end{itemize}

\section{Related Work}
In this section, we briefly review existing works in backdoor attack and defense that are most relevant to our work.

\subsection{Backdoor Attack}

A backdoor attack implants a backdoor trigger into the victim model by injecting the trigger pattern into a small proportion of the training data. It preserves the model's performance on benign (clean) inputs but can manipulate the model to constantly predicts the attacker-specified backdoor class whenever the trigger pattern appears in a test input.

\subsubsection{BadNets}
BadNets \cite{gu2017badnets} is the first backdoor attack that was designed for image classification models. With an attacker-specified backdoor label $y_{t}$, BadNets first stamps a pre-designed backdoor trigger onto a benign image $\bm{x}$ to generate a poisoned sample $\bm{x'}$ and changes its ground-truth label to $y_{t}$. It then trains a backdoored model on the poisoned training data of which a small portion of the samples are poisoned following the above procedure.
At inference time, the attacked model performs well on benign test samples, yet consistently predicts the backdoor label $y_{t}$ for any test samples with the trigger pattern attached. Using a simple checkerboard pattern, BadNets can achieve an attack success rate (\textit{i.e.,} the ratio of poisoned test samples that are predicted as the backdoor class) of 99\% on MNIST dataset by poisoning only 10\% of the training data. In Section \ref{sec:baseline}, we will propose a simple BadNets-equivalent baseline attack for time series.

\subsubsection{Invisible patterns}

Following BadNets, a number of backdoor attacks have been proposed in computer vision applications with images. \cite{chen2017targeted} first discussed the stealthiness of backdoor attacks in regard to the invisibility requirement of trigger patterns. They suggested that poisoned images should be indistinguishable from their benign counter-part to evade human inspection. Accordingly, they proposed a blending strategy that generates poisoned images by blending the backdoor trigger with benign images, instead of direct stamping. A small-amplitude random noise is then added to further reduce the risk of being detected. After \cite{chen2017targeted}, a series of works have been proposed to generate invisible trigger patterns, which include \cite{liu2020reflection,turner2019label,cheng2020deep,bagdasaryan2021blind}. All these works were proposed for images. In \cite{zhao2020clean}, a video backdoor attack was proposed against video recognition models. It leverages universal adversarial perturbations to tackle the higher dimension challenge posed by videos. Backdoor attacks have also been extended to other vision tasks such as crowd counting \cite{sun2022backdoor} and visual object tracking \cite{li2022few}. In this work, we will also limit our patterns to be invisible, but address the lower dimension challenge posed by time series.

\subsubsection{Sample-specific patterns}
The above-mentioned backdoor attacks all use a fixed pattern at a fixed location of the image as the trigger pattern, which could potentially be defended and removed easily. To address this problem, \cite{nguyen2020input} proposed an input-aware backdoor attack to generate different backdoor trigger patterns for different input samples, enforcing each trigger pattern to be only functional for one particular input sample. Similarly, \cite{doan2021lira} proposed LIRA that can generate the optimal trigger while successfully poisoning the classifier in computer vision applications. Most recently, inspired by DNN-based image steganography, \cite{li2021invisible} proposed another sample-specific backdoor attack via encoding an attacker-specified string into benign images, which generates sample-specific additive noises as backdoor triggers. For stealthiness, in this paper, we are also interested in sample-specific backdoor attacks.

\subsubsection{Time series attacks}
\cite{wang2020backdoor} studied backdoor attacks against transfer learning on both image and time series data. Particularly, they proposed to manipulate the pre-trained teacher models to generate customized student models that make the wrong predictions. However, the times series (e.g. ECG signals) were transformed into 1D or 2D images and attacked based on image backdoor attacks.
\cite{ning2022trojanflow} proposed TrojanFlow to perform backdoor attacks against network traffic classification models. By poisoning the traffic flow in a stealthy manner, this attack achieves high ASRs on both flow-based and payload-based classifiers. More recently, \cite{ding2022towards} proposed a backdoor attack TimeTrojan on time series classifiers. TimeTrojan proposes a multi-objective optimization framework to help the model learn a strong correlation between the trigger pattern and the target class.

There are also backdoor attacks that are not realized via data poisoning but direct modifications of the model parameters \cite{dumford2020backdooring, garg2020can} or structures \cite{tang2020embarrassingly, li2021deeppayload, qi2021subnet}. These attacks are independent of poisoning-based attacks and can be performed even after a poisoning-based attack. In this paper, we will focus on data poisoning-based attacks and leave other types of manipulations to future work. Particularly, we propose a generative approach for time series backdoor attacks that produce both realistic and effective sample-wise backdoor triggers.

\subsection{Backdoor Defense}

A large number of defense methods have been proposed to mitigate the backdoor threat via detection or purification. 
Neural Cleanse \cite{wang2019neural} presented the first solution to detect poisoned models. It first computes potential trigger patterns for each class that could convert any clean image to that class. Then, it detects among these candidates and selects the abnormally smaller ones as backdoor indicators. Following \cite{wang2019neural}, improved detection techniques were introduced in \cite{chen2019deepinspect,harikumar2020scalable}.
Fine-Pruning \cite{liu2018fine} purifies a backdoored model by eliminating neurons that are dormant on clean inputs.
Knowledge distillation (KD) \cite{hinton2015distilling} techniques were also leveraged in \cite{yoshida2020disabling} and \cite{li2021neural} to remove backdoors from infected DNNs. However, applying Fine-Pruning and KD could degrade the clean accuracy when only limited clean data are available \cite{chen2021refit}.  \cite{xu2020defending} proposed to remove neurons with high activation values from the final convolutional layer. More recently, a robust training strategy was proposed in \cite{li2021anti} to train backdoor-free models on the poisoned dataset. \cite{wu2021adversarial} introduces an effective backdoor removal method Adversarial Neuron Pruning (ANP) to prune adversarially sensitive neurons to purify the model.

In this paper, we present a novel generative time series backdoor attack that can produce backdoor samples that are not detectable by either human inspectors or strong backdoor defenses.



\section{Time Series Backdoor Attack}

In this section, we introduce our proposed \emph{Time Series Backdoor Attack (TSBA)} in the context of time series classification (TSC).
We first define our threat model and overview the attack pipeline, then introduce the details of the proposed trigger generator and its training procedure. Finally, we introduce how to train a universal trigger generator to craft sample-specific backdoor triggers for any type of time series.

\subsection{Problem Formulation}
\label{sec:prob}

Let $\gD=\{(\vx_{i}, y_{i})\}_{i=1}^{N}$ denote the benign training set of $N$ \textit{i.i.d.} samples, where each $\vx_{i}$ represents one time series sample and $y_{i}$ is its corresponding ground-truth label. A classification model $f$ learns the function $f: \gX \rightarrow \gY$ that maps the input space to the label space. This is typically done by minimizing the model's classification error on $\gD$ as follows:
\begin{equation}
    \min_{\theta} \E_{(\vx,y) \in \gD} \Ls_{CE}(f(\vx), y),
\end{equation}
where $\mathcal{L}_{CE}$ is the commonly used cross-entropy loss and $\theta$ are the model parameters.

A backdoor adversary poisons the training data $\gD$ into a poisoned dataset $\gD'$ with a trigger pattern $\vp$ such that the model trained on $\gD'$ will become a backdoored model $f'$. With the trigger pattern $\vp$, a poisoned sample can be crafted by $\vx'=\vx \odot \vp$ where $\odot$ represents any stamping method, such as addition, subtraction, or replacement. 
The adversary aims to achieve two goals with the backdoored model $f'$. First, the model should predict the correct label for benign inputs, \textit{i.e.,} $f'(\vx) = y$ for any test input $\vx \in \gD_{test}$. Second, the model should predict the backdoor class $y_t$ for any stamped test input with the trigger pattern, \textit{i.e.,} $f'(\vx') = y_t$ for any backdoored test input $\vx'$.


\subsection{Threat Model}
\label{sec:threat_model}
Existing threat models from the adversary's perspective can be categorized into three main categories: 1) \emph{training manipulation} where the adversary not only poisons the training data but also controls the training procedure \cite{cheng2020deep,nguyen2020input,yao2019latent}; 2) \emph{data poisoning} where the adversary can only poison the training data with the trigger pattern \cite{gu2017badnets,chen2017targeted}; and 3) \emph{post-training injection} where the adversary does not poison the training data nor control the training procedure, but directly modifies the parameters of a cleanly-trained model \cite{liu2017trojaning}.
Unlike most image backdoor attacks that only consider one of the above three threat models, in this paper we design and evaluate our \emph{Time Series Backdoor Attack (TSBA)} under two different threat models that fall into categories 1) and 2) stated above, denoted as \textit{TSBA-A} and \textit{TSBA-B}:

\subsubsection*{\textbf{TSBA-A}}
This threat model simulates real-world scenarios where the victim users directly download and deploy pre-trained DNNs models from untrusted sources. Under this threat model, the adversary has full access to the training data $\gD$ with complete control over the training procedure. 
Accordingly, the adversary purposely trains a backdoored model $f'$ with the poisoned training data $\gD'$ which contains a small portion of poisoned samples by sample-wise trigger pattern $\vp$. 
After downloading the backdoored model $f'$, the victim user is expected to inspect the model's performance on its own validation data $D_{val}$ which is unknown to the adversary.

\subsubsection*{\textbf{TSBA-B}}
This threat model assumes the victim users have full control over the training procedure but accidentally collected a few poisoned samples into its training set. The adversary may have knowledge of the network architecture adopted by the victim, and can poison a small number of training samples.
The adversary can thus add sample-wise trigger patterns to a few samples and then injects the samples into the training set. The victim trains a backdoored model $f'$ on the poisoned dataset $\gD'$. Meanwhile, the victim may inspect the poisoned dataset $\gD'$ via either visual inspection or certain backdoor defense techniques to remove the backdoored samples.

\subsection{Proposed Attack Pipeline}

\begin{figure}[h]
\centering
\includegraphics[width=\columnwidth]{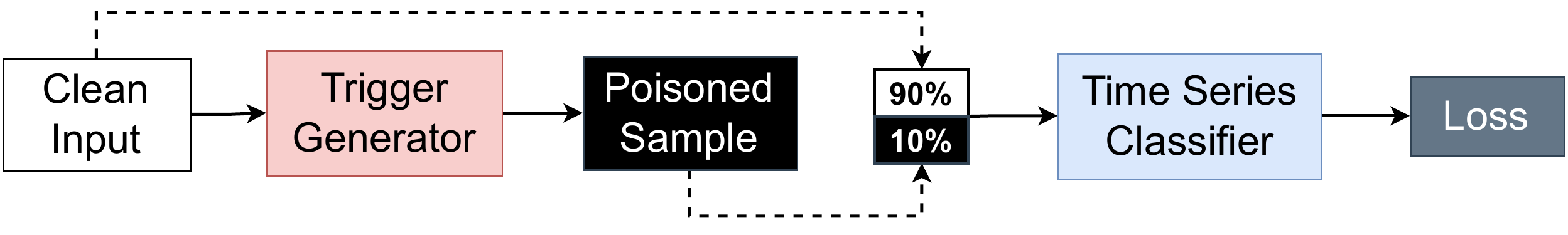} \\ \vspace{1ex}
\includegraphics[width=\columnwidth]{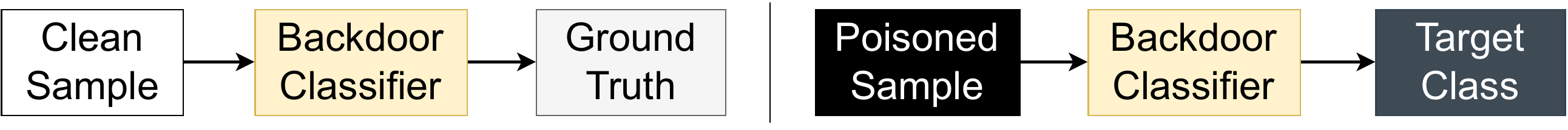}
\caption{Overview of the proposed TSBA attack. \emph{Top}: training the TSBA trigger pattern generator; \emph{Bottom}: inference with the backdoored model on clean vs poisoned samples. }
\label{fig:pipeline}
\end{figure}

As illustrated in Figure \ref{fig:pipeline}, our proposed TSBA attack trains a trigger generator network to generate sample-specific backdoor trigger patterns for poisoned samples. 
The adversary randomly selects 10\% of training data for backdoor poisoning, and trains a time series classifier on both the poisoned and clean training samples. 
To achieve both a high attack success rate and high clean accuracy, the two components of TSBA,  namely the trigger generator and the classifier, are trained iteratively via specific procedures described in Section \ref{sec:model_training}. 
At inference time, the backdoored classifier will predict the adversary-specified class (backdoor class $y_t$) for poisoned samples, while recognizing correct classes for clean samples.

Under the \textit{TSBA-A} threat model, the adversary directly releases the backdoored classifier to the victim user, while privately keeping the trained trigger generator to perform backdoor attacks at inference time, \textit{i.e.,} as our TSBA is a sample-specific attack, the adversary will need the trigger generator to produce backdoor samples at inference time.

Under the \textit{TSBA-B} threat model, the adversary will leverage the trigger generator to perform data poisoning, \textit{i.e.,} poisoning a small proportion (e.g. 10\%) of the training samples using the trigger generator. The victim user then trains a classifier on the poisoned training set following a typical model training procedure. The victim user will use the trained classifier to perform inference on either clean or backdoored test samples. The inference procedure is the same as in \textit{TSBA-A}.

\begin{algorithm}[h!]
\DontPrintSemicolon
  Let $f$ be the classifier, $g$ be the trigger generator
  
  Let $\text{clip}_{\xi}(\vx \odot \vk)$ limit $\vx$ bounded by $[\vk-\xi, \vk+\xi]$
  
  Let $G_{\xi}(\vx) = \text{clip}_{\xi}(g(\vx) \odot \vx)$

  Given a target class $y_{t}$, a training dataset $\gD$, the backdoor poison rate $\gamma$
  
  $T_{clean}$: clean training epochs; $T_{backdoor}$: backdoor training epochs

  Initialize $f, g$

\# warm start $f$

\For{$i$ in $range(T_{clean})$}
{
	\For{$(\vx, y)$ in $\gD$}
	{
	$f \leftarrow \argmin_{f}\mathcal{L}_{CE}(f(\vx), y)$
	}
}

\# simultaneous training of $g$ and $f$

$\gD_{p} \leftarrow \text{random\_sample}_{\gamma}(\gD)$

\For{$i$ in $range(T_{backdoor})$}
{

	\For{$(\vx', y_{t})$ in $\gD_{p}$}
	{
		$g$ $\leftarrow \argmin_{g} \mathcal{L}_{CE}(f(G_{\xi}(\vx')), y_{t})$
	}

        \# update the poisoned training data
        
	$\gD' \leftarrow \gD \cup  \{G_{\xi}(\gD_{p}), y_{t}\}$

	\For{$(\vx', y')$ in $\gD'$}
	{

		$f \leftarrow \argmin_{f}\mathcal{L}_{CE}(f(\vx'), y')$
	}
	
\Indm
    
}
   \Return $f$, $g$

\caption{Training Procedure of TSBA}
\label{algo:generator}
\end{algorithm}

The core component of TSBA is the trigger generator which uses a simple DNN architecture suited for both univariate and multivariate time series. 
The detailed model structure is shown in Table \ref{table:noise_gen}. The trigger generator takes a time series sample as an input and generates a sample-specific trigger pattern for the sample. Test the trigger pattern is of the same size as the original sample. Then, the generated pattern is added to the original sample to craft a poisoned sample.
Given that the trigger patterns are dynamic (sample-specific) and stealthy (ensured by the generator), it is difficult for the victim user to identify which samples are backdoored \cite{nguyen2020input,li2021invisible}. Moreover, even if the victim user has detected the trigger patterns for several samples, they could not remove the trigger pattern for other samples without the trigger generator. This is in sharp contrast to fixed-pattern based backdoor attacks, where the fixed pattern can be easily removed once detected.

\subsection{Training the Trigger Generator}
\label{sec:model_training}

The training procedure of the trigger generator is described in Algorithm \ref{algo:generator}.
which iteratively trains the trigger generator  $g$ and the backdoored classifier $f$ to achieve both a high attack success rate and high clean accuracy.

\begin{figure}[h!]
\includegraphics[width=\columnwidth]{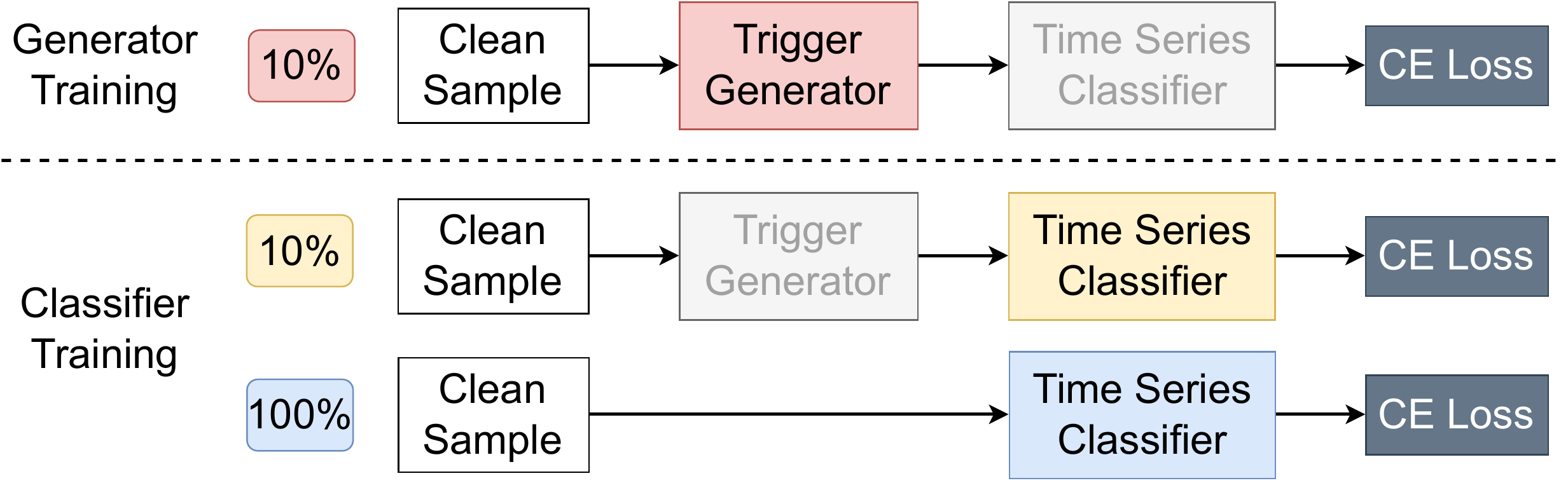}
\centering
\caption{The training procedure of the trigger generator $g$ (top) and the backdoor classifier $f$ (bottom)}
\label{fig:pipeline_train}
\end{figure}

To address the cold start problem of generator training, we first pre-train the time series classifier $f$ for $T_{clean}$ epochs on all clean samples from $\gD$ until it has a steady drop in the cross-entropy loss $\Ls_{CE}$. This corresponds to lines 7-11 in Algorithm \ref{algo:generator}.
After this pre-training, we train the trigger generator $g$ and the partially trained classifier $f$ simultaneously for $T_{backdoor}$ epochs.
Both networks are progressively updated in each iteration following a similar process: 1) training $g$ on the poisoned samples (initialized to be randomly sampled clean samples at line 13) to minimize the classification loss with respect to the backdoor class $y_t$ (line 15-17); 2) generating poisoned dataset $\gD'$ using $g$ (line 18); and 3) training $f$ on the poisoned dataset $\gD'$ with the poisoned samples are relabeled to $y_t$ (line 19-21).
Note that during the entire process, the backdoor trigger pattern is clipped to be  within 10\% of the signal amplitude, \textit{i.e.,} $0.1*(\vx_{max} - \vx_{min})$, to strengthen stealthiness (line 2).
This process is further illustrated in Figure \ref{fig:pipeline_train}. This training procedure encourages the trigger generator $g$ to explore the most effective patterns that can alter $f$'s predictions towards the target class $y_t$.

Unlike the image backdoor training where the backdoor samples are generated before model training, we refresh the sample-specific trigger for each backdoor sample using the updated trigger generator $g$. Likewise, the time series classifier $f$ is backdoor trained to minimize the CE loss on the partially-poisoned training data. Thus, it enables the classifier to recognize the backdoor pattern induced by the trigger generator, while maintaining clean accuracy with clean training samples.

\noindent\textbf{How TSBA works?} In TSBA, the generator is designed to progressively generate stronger (and more realistic) trigger patterns, while the classifier simultaneously learns the correlation between the trigger patterns and the target class. This design is motivated by our observation in Section \ref{sec:understanding} that simple trigger patterns cannot be easily learned into the model. I.e., establishing the backdoor correlation is relatively hard in time series models. The clean signals tend to overwhelm the backdoor noise during the training process. As such, the generator and the classifier need to learn together to explore stronger triggers. Beyond time series backdoors, our design could also be useful for scenarios where backdoor triggers are generally hard to inject.

\subsection{Training a Universal Trigger Generator}
\label{sec:universal}

The above generator needs to be re-trained or fine-tuned for a new type of time series.
Here, we further train a universal trigger generator to generate sample-specific triggers for any type of time series without the need to fine-tune the generator for unseen datasets.
The universal generator has a similar architecture as the dataset-specific trigger generator used in the above \textit{TSBA} algorithm, but with one additional convolutional layer and revised parameters to provide more generalization capacities for multiple time series datasets. The detailed model architecture can be found in Table \ref{table:uni_gen}.


\begin{algorithm}[h]
\DontPrintSemicolon
  Let $g$ be the universal trigger generator

  Given a set of distinct time series datasets $\mathcal{S}=\{\mathcal{S}_{1}, \mathcal{S}_{2}, ..., \mathcal{S}_{n}\}$
  
  Let $\mathcal{C} = \{1, 2, ..., n\}$ be a class set
  

  Initialize $g$

\For{$i$ in $range(T_u)$}
{	
	$\mathcal{S'} \leftarrow \text{random\_sample}(\mathcal{S})$

	$y_t \leftarrow \text{random\_select}(\mathcal{C})$

	$g \leftarrow$  TSBA($g, \mathcal{S'}, y_t$)  \ \ \ \ $\triangleright$ \textit{following} \textbf{Algorithm \ref{algo:generator}}
}
   \Return $g$

\caption{Universal Trigger Generator Training}
\label{algo:uni_generator}
\end{algorithm}

As described in Algorithm \ref{algo:uni_generator}, it generally follows the training procedure of \textit{TSBA}  with several extra steps. The generator is trained with samples from a combination of time series datasets which are merged with all class labels re-organized. 
In each iteration, the generator is refreshed with randomly selected training samples and target classes. Accordingly, the generator is optimized to create the trigger pattern according to the style of the time series input to perturb samples from unseen datasets. Thus, it simplifies and automates the procedure of trigger pattern selection by removing the necessity for manual selection or training a dataset-specific trigger generator. More detailed training setup can be found in Section \ref{sec:experiment}.

\section{Experiments}\label{sec:experiment}

In this section, we evaluate the performance of our proposed TSBA and the universal trigger generator. We first introduce three simple baseline attacks and describe our experimental setup. We then discuss the experimental results and show the differences between image vs. time series backdoor attacks. Following that, we demonstrate the stealthiness of the backdoored samples generated by TSBA, as well as its resistance to backdoor defenses. 

\begin{figure*}[ht]
\includegraphics[width=2.0\columnwidth]{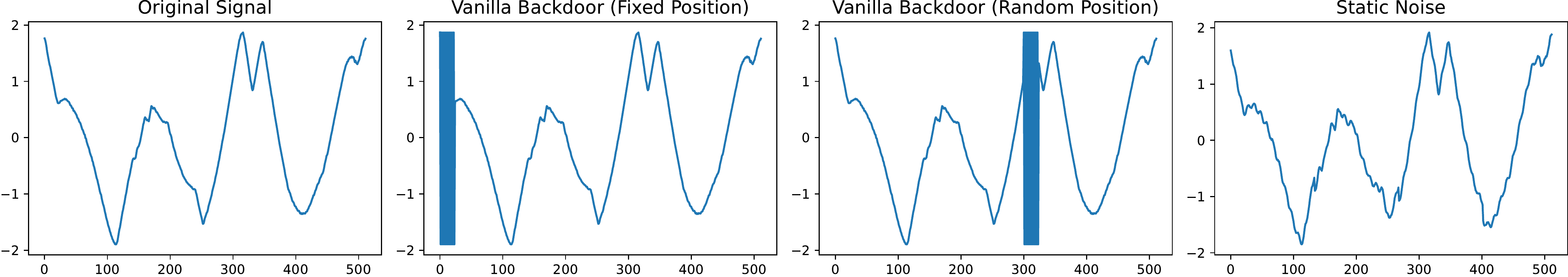}
\vspace{1ex}
\centering
\caption{Waveforms of the original and attacked signals by 3 baseline attacks. The clean signal is taken from ($D_{1}$) BirdChicken.}
\label{fig:comp_pattern}
\end{figure*}

\begin{table*}[h!]
\vspace{3ex}
\renewcommand{\arraystretch}{1.05}
\centering
\caption{Experiment results of poison-label backdoor attacks on 13 datasets in terms of clean accuracy (\textbf{CA}) and attack success rate (\textbf{ASR}). Both \textit{TSBA-A} and \textit{TSBA-B} threat models are tested.
$D_{1}$ to $D_{8}$ are univariate datasets from the UCR Archive, while $D_{9}$ to $D_{13}$ are multivariate datasets from the MTS Archive. The accuracy of clean model (without backdoor training) is provided in the $3^{rd}$ column (\textbf{Clean}).}
\begin{tabular}{lcccccccccccc}
\toprule
\multicolumn{1}{l}{}  \makecell{\multirow{2}{*}{\vspace{-1.5ex}\textbf{Dataset}}}  &   \multirow{2}{*}{\vspace{-1.5ex}\textbf{Classifier}}   &  \multirow{2}{*}{\vspace{-1.5ex}\textbf{Clean}}   & \multicolumn{2}{c}{Vanilla (Fixed)}      & \multicolumn{2}{c}{Vanilla (Random)}             & \multicolumn{2}{c}{Static Noise}                    & \multicolumn{2}{c}{\textbf{TSBA-A (Ours)}}              & \multicolumn{2}{c}{\textbf{TSBA-B (Ours)}}                  \\

 \cmidrule{4-13}
    &    &   & \textbf{CA} & \multicolumn{1}{c}{\textbf{ASR}} & \textbf{CA} & \multicolumn{1}{c}{\textbf{ASR}} & \textbf{CA} & \multicolumn{1}{c}{\textbf{ASR}} & \textbf{CA} & \multicolumn{1}{c}{\textbf{ASR}} & \textbf{CA} & \multicolumn{1}{c}{\textbf{ASR}}  \\ \midrule

\multirow{3}{*}{\begin{tabular}[c]{@{}c@{}} ($D_{1}$) BirdChicken\end{tabular}}  &  
   TCN       & 96.0\%    & 88.0\%    & 52.0\%    & 86.0\%    & 56.0\%    & 90.0\%    & 79.0\%    & 94.0\%    & \textbf{100.0\%}    & 92.0\%    & 94.0\%  \\
& ResNet  & 92.0\%    & 86.0\%    & 48.0\%    & 83.0\%    & 49.0\%    & 85.0\%    & 80.0\%    & 90.0\%    & \textbf{100.0\%}    & 86.0\%    & 94.0\%  \\
& LSTM     & 94.0\%    & 85.0\%    & 49.0\%    & 83.0\%    & 48.0\%    & 88.0\%    & 72.0\%    & 91.0\%    & \textbf{100.0\%}    & 88.0\%    & 92.0\%  \\ \midrule

\multirow{3}{*}{\begin{tabular}[c]{@{}c@{}} ($D_{2}$) ECG5000\end{tabular}}  &  
   TCN       & 94.6\%    & 88.2\%    & 56.1\%    & 85.0\%    & 55.9\%    & 89.6\%    & 76.2\%    & 92.7\%    & \textbf{100.0\%}    & 89.5\%    & 89.0\%  \\
& ResNet  & 93.9\%    & 88.9\%    & 55.6\%    & 85.2\%    & 55.8\%    & 89.8\%    & 75.9\%    & 92.3\%    & \textbf{100.0\%}    & 89.0\%    & 86.2\%  \\
& LSTM     & 94.1\%    & 86.9\%    & 55.2\%    & 85.1\%    & 54.7\%    & 88.5\%    & 74.7\%    & 92.0\%    & \textbf{99.5\%}    & 89.1\%    & 84.1\%  \\ \midrule

\multirow{3}{*}{\begin{tabular}[c]{@{}c@{}} ($D_{3}$) Earthquakes\end{tabular}}  &  
   TCN       & 72.5\%    & 66.9\%    & 51.7\%    & 63.9\%    & 55.8\%    & 66.6\%    & 72.1\%    & 71.4\%    & \textbf{100.0\%}    & 66.9\%    & 81.1\%  \\
& ResNet  & 71.7\%    & 66.9\%    & 53.9\%    & 65.8\%    & 54.0\%    & 66.5\%    & 74.6\%    & 70.3\%    & \textbf{99.7\%}    & 66.3\%    & 78.4\%  \\
& LSTM     & 72.2\%    & 64.6\%    & 53.3\%    & 64.8\%    & 56.9\%    & 68.1\%    & 70.4\%    & 68.8\%    & \textbf{100.0\%}    & 66.0\%    & 76.8\%  \\ \midrule

\multirow{3}{*}{\begin{tabular}[c]{@{}c@{}} ($D_{4}$) ElectricDevices\end{tabular}}  &  
   TCN       & 72.3\%    & 65.9\%    & 47.1\%    & 63.6\%    & 49.7\%    & 66.9\%    & 76.1\%    & 70.7\%    & \textbf{100.0\%}    & 67.1\%    & 82.5\%  \\
& ResNet  & 73.4\%    & 66.3\%    & 47.8\%    & 66.4\%    & 48.9\%    & 68.3\%    & 75.8\%    & 71.8\%    & \textbf{99.8\%}    & 68.6\%    & 81.6\%  \\
& LSTM     & 70.8\%    & 64.9\%    & 46.5\%    & 63.7\%    & 44.2\%    & 64.8\%    & 71.9\%    & 68.9\%    & \textbf{99.4\%}    & 65.9\%    & 81.9\%  \\ \midrule

\multirow{3}{*}{\begin{tabular}[c]{@{}c@{}} ($D_{5}$) Haptics\end{tabular}}  &  
   TCN       & 50.2\%    & 46.4\%    & 44.5\%    & 45.0\%    & 49.1\%    & 46.1\%    & 69.0\%    & 49.3\%    & \textbf{99.0\%}    & 46.5\%    & 86.7\%  \\
& ResNet  & 51.6\%    & 48.5\%    & 45.7\%    & 47.2\%    & 46.7\%    & 47.7\%    & 66.8\%    & 50.8\%    & \textbf{99.7\%}    & 48.4\%    & 88.2\%  \\
& LSTM     & 54.0\%    & 48.1\%    & 46.3\%    & 47.0\%    & 48.6\%    & 48.9\%    & 67.2\%    & 52.8\%    & \textbf{99.6\%}    & 49.6\%    & 88.6\%  \\ \midrule

\multirow{3}{*}{\begin{tabular}[c]{@{}c@{}} ($D_{6}$) PowerCons\end{tabular}}  &  
   TCN       & 88.2\%    & 79.7\%    & 59.1\%    & 76.1\%    &62.0\%    & 80.6\%    & 76.4\%    & 84.6\%    & \textbf{100.0\%}    & 81.7\%    & 79.1\%  \\
& ResNet  & 89.7\%    & 82.6\%    & 58.2\%    & 79.1\%    & 61.2\%    & 82.9\%    & 75.2\%    & 85.7\%    & \textbf{100.0\%}    & 84.3\%    & 78.2\%  \\
& LSTM     & 86.4\%    & 76.8\%    & 56.7\%    & 72.7\%    & 61.7\%    & 78.2\%    & 73.9\%    & 81.3\%    & \textbf{100.0\%}    & 79.3\%    & 75.6\%  \\ \midrule

\multirow{3}{*}{\begin{tabular}[c]{@{}c@{}} ($D_{7}$) ShapeletSim\end{tabular}}  &  
   TCN       & 72.4\%    & 64.0\%    & 56.9\%    & 61.3\%    & 58.2\%    & 64.3\%    & 77.9\%    & 69.0\%    & \textbf{100.0\%}    & 66.5\%    & 84.7\%  \\
& ResNet  & 77.9\%    & 70.8\%    & 57.0\%    & 65.2\%    & 56.7\%    & 69.9\%    & 79.4\%    & 73.8\%    & \textbf{100.0\%}   & 71.5\%    & 85.2\%  \\
& LSTM     & 68.5\%    & 62.1\%    & 55.3\%    & 55.3\%    & 56.9\%    & 61.8\%    & 76.2\%    & 63.7\%    & \textbf{99.2\%}   & 60.4\%    & 84.4\%  \\ \midrule

\multirow{3}{*}{\begin{tabular}[c]{@{}c@{}} ($D_{8}$) Wine\end{tabular}}  &  
   TCN       & 59.6\%    & 55.5\%    & 52.6\%    & 52.5\%    & 54.1\%    & 56.2\%    & 74.0\%    & 57.5\%    & \textbf{98.4\%}    & 56.5\%    & 80.7\%  \\
& ResNet  & 74.5\%    & 66.1\%    & 50.1\%    & 66.3\%    & 52.7\%    & 71.3\%    & 76.2\%    & 71.6\%    & \textbf{96.9\%}    & 71.9\%    & 76.9\%  \\
& LSTM     & 66.8\%    & 59.9\%    & 48.5\%    & 58.9\%    & 53.4\%    & 64.0\%    & 73.7\%    & 64.4\%    & \textbf{95.8\%}    & 64.1\%    & 78.2\%  \\ \midrule \midrule

\multirow{3}{*}{\begin{tabular}[c]{@{}c@{}} ($D_{9}$) ArabicDigits\end{tabular}}  &  
   TCN       & 99.4\%    & 92.8\%    & 52.4\%    & 90.6\%    & 56.8\%    & 93.5\%    & 68.7\%    & 97.1\%    & \textbf{96.1\%}    & 94.2\%    & 77.4\%  \\
& ResNet  & 99.6\%    & 92.5\%    & 56.7\%    & 90.2\%    & 58.4\%    & 94.2\%    & 65.2\%    & 96.6\%    & \textbf{98.2\%}    & 94.2\%    & 78.9\%  \\
& LSTM     & 94.2\%    & 89.2\%    & 55.3\%    & 86.3\%    & 56.1\%    & 88.1\%    & 64.9\%    & 91.8\%    & \textbf{94.5\%}    & 88.2\%    & 75.0\%  \\ \midrule

\multirow{3}{*}{\begin{tabular}[c]{@{}c@{}} ($D_{10}$) ECG\end{tabular}}  &  
   TCN       & 87.4\%    & 82.5\%    & 59.1\%    & 80.5\%    & 59.0\%    & 82.3\%    & 79.2\%    & 85.5\%    & \textbf{100.0\%}    & 82.8\%    & 81.6\%  \\
& ResNet  & 86.2\%    & 82.0\%    & 62.4\%    & 79.0\%    & 61.2\%    & 81.0\%    & 80.1\%    & 84.8\%    & \textbf{100.0\%}    & 81.5\%    & 83.7\%  \\
& LSTM     & 86.8\%    & 83.7\%    & 58.6\%    & 82.0\%    & 60.5\%    & 81.4\%    & 77.4\%    & 84.7\%    & \textbf{100.0\%}    & 83.0\%    & 80.5\%  \\ \midrule

\multirow{3}{*}{\begin{tabular}[c]{@{}c@{}} ($D_{11}$) KickvsPunch\end{tabular}}  &  
   TCN       & 54.0\%    & 51.1\%    & 55.1\%    & 48.8\%    & 55.4\%    & 50.9\%    & 74.1\%    & 53.2\%    & \textbf{98.4\%}    & 51.0\%    & 77.1\%  \\
& ResNet  & 51.3\%    & 47.9\%    & 54.3\%    & 46.4\%    & 56.9\%    & 48.5\%    & 75.9\%    & 50.4\%    & \textbf{97.1\%}    & 48.7\%    & 75.8\%  \\
& LSTM     & 51.1\%    & 49.3\%    & 52.7\%    & 47.5\%    & 53.1\%    & 49.1\%    & 76.0\%    & 50.0\%    & \textbf{97.6\%}    & 48.8\%    & 75.4\%  \\ \midrule

\multirow{3}{*}{\begin{tabular}[c]{@{}c@{}} ($D_{12}$) NetFlow\end{tabular}}  &  
   TCN       & 89.4\%    & 83.7\%    & 48.9\%    & 79.7\%    & 49.6\%    & 83.7\%    & 79.4\%    & 86.3\%    & \textbf{100.0\%}    & 84.5\%    & 82.4\%  \\
& ResNet  & 77.5\%    & 72.6\%    & 51.1\%    & 69.4\%    & 50.8\%    & 72.3\%    & 80.6\%    & 75.1\%    & \textbf{100.0\%}    & 72.9\%    & 84.5\%  \\
& LSTM     & 88.6\%    & 82.2\%    & 50.4\%    & 80.4\%    & 50.1\%    & 82.1\%    & 78.5\%    & 85.4\%    & \textbf{100.0\%}    & 83.4\%    & 84.9\%  \\ \midrule

\multirow{3}{*}{\begin{tabular}[c]{@{}c@{}} ($D_{13}$) UWave\end{tabular}}  &  
   TCN       & 93.4\%    & 87.0\%    & 43.7\%    & 83.2\%    & 46.7\%    & 87.5\%    & 66.7\%    & 90.4\%    & \textbf{98.0\%}    & 87.3\%    & 72.5\%  \\
& ResNet  & 92.2\%    & 85.5\%    & 46.2\%    & 80.8\%    & 48.9\%    & 86.0\%    & 68.5\%    & 88.8\%    & \textbf{99.1\%}    & 86.5\%    & 76.7\%  \\
& LSTM     & 84.1\%    & 78.9\%    & 42.5\%    & 74.3\%    & 46.5\%    & 79.0\%    & 64.3\%    & 81.3\%    & \textbf{96.4\%}    & 78.7\%    & 69.4\%  \\

\bottomrule
\end{tabular}
\label{table:exp_backdoor}
\end{table*}

\subsection{Simple Baseline Attacks}
\label{sec:baseline}

In this work, we also propose three image-equivalent methods of time series backdoor attacks. Due to the lack of prior research on time series backdoor attacks, we use them as the baseline attacks. The attacked waveforms by the 3 baseline attacks are illustrated in Figure \ref{fig:comp_pattern}.
The first two baselines are the two versions of a \textit{vanilla backdoor} attack: 1) adding a fixed pattern at the beginning of the time series, or 2) covering the peaks (or troughs) in the time series by a fixed pattern. Here we set the fixed pattern to be the min/max value of the entire time series, as they are the most salient signals.
Both baseline attacks randomly choose and alter 5\% of the time series dimensions alternatively to their maximum and minimum signal values.

The third baseline attack, named \textit{static noise}, uses static powerline noise as the trigger pattern. This idea was motivated by the observation in recent research that powerline noises caused by the signal capture device commonly exist in Electrocardiogram (ECG) signals \cite{gilani2018power, bahaz2018efficient}. The noise is tiled (repeated) to the same length as the original time series and applied to all time-dependent dimensions. Also, it is standardized with its size set to be 10\% of the amplitude ($\vx_{max} - \vx_{min}$), which is of the same amplitude as the patterns generated by our TSBA.

\subsection{Experimental Setup}
\label{sec:exp_setup}

\begin{figure*}[ht]
\centering
\begin{tabular}{c|c}
\includegraphics[width=.98\columnwidth]{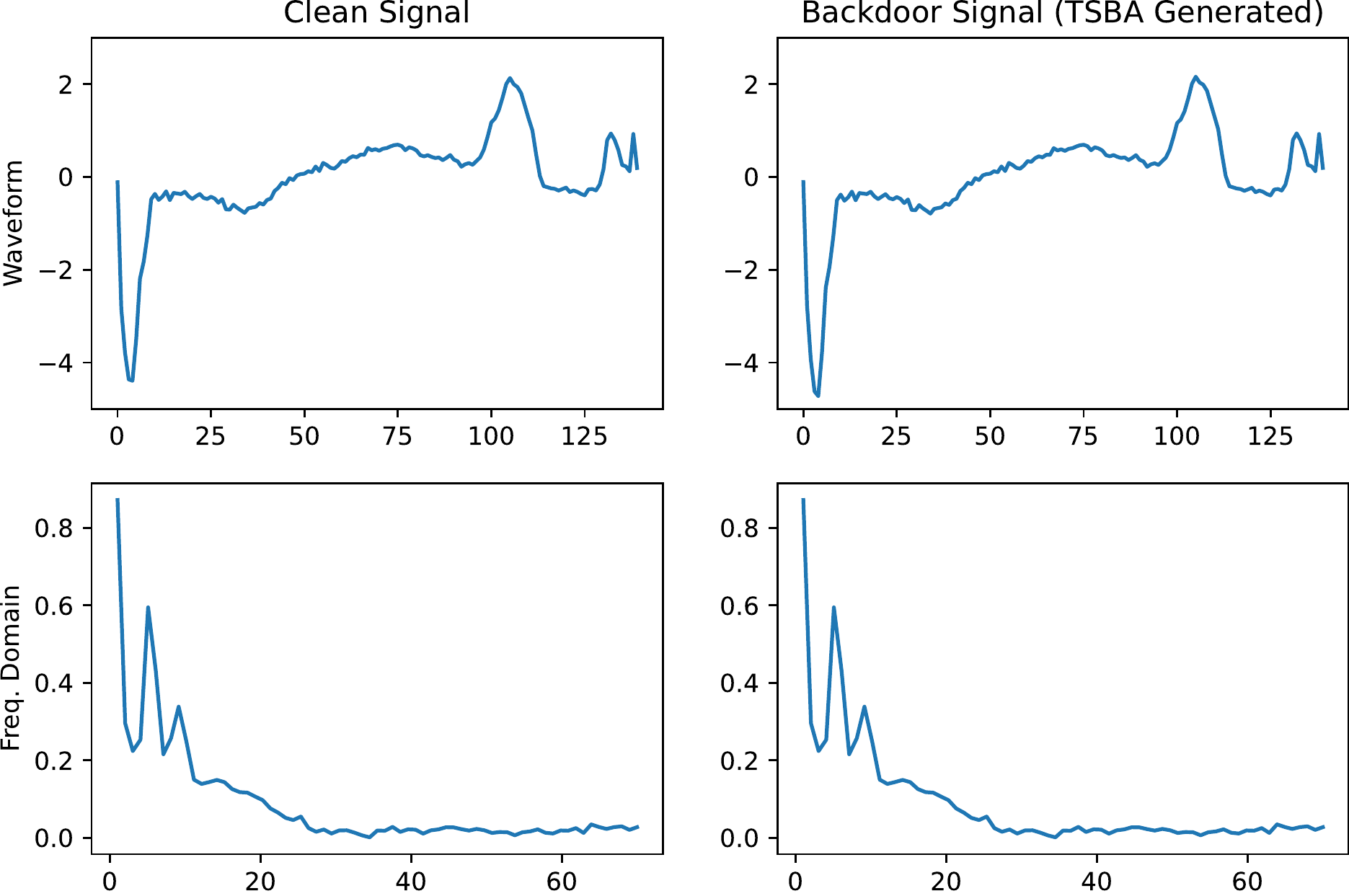} & \includegraphics[width=.98\columnwidth]{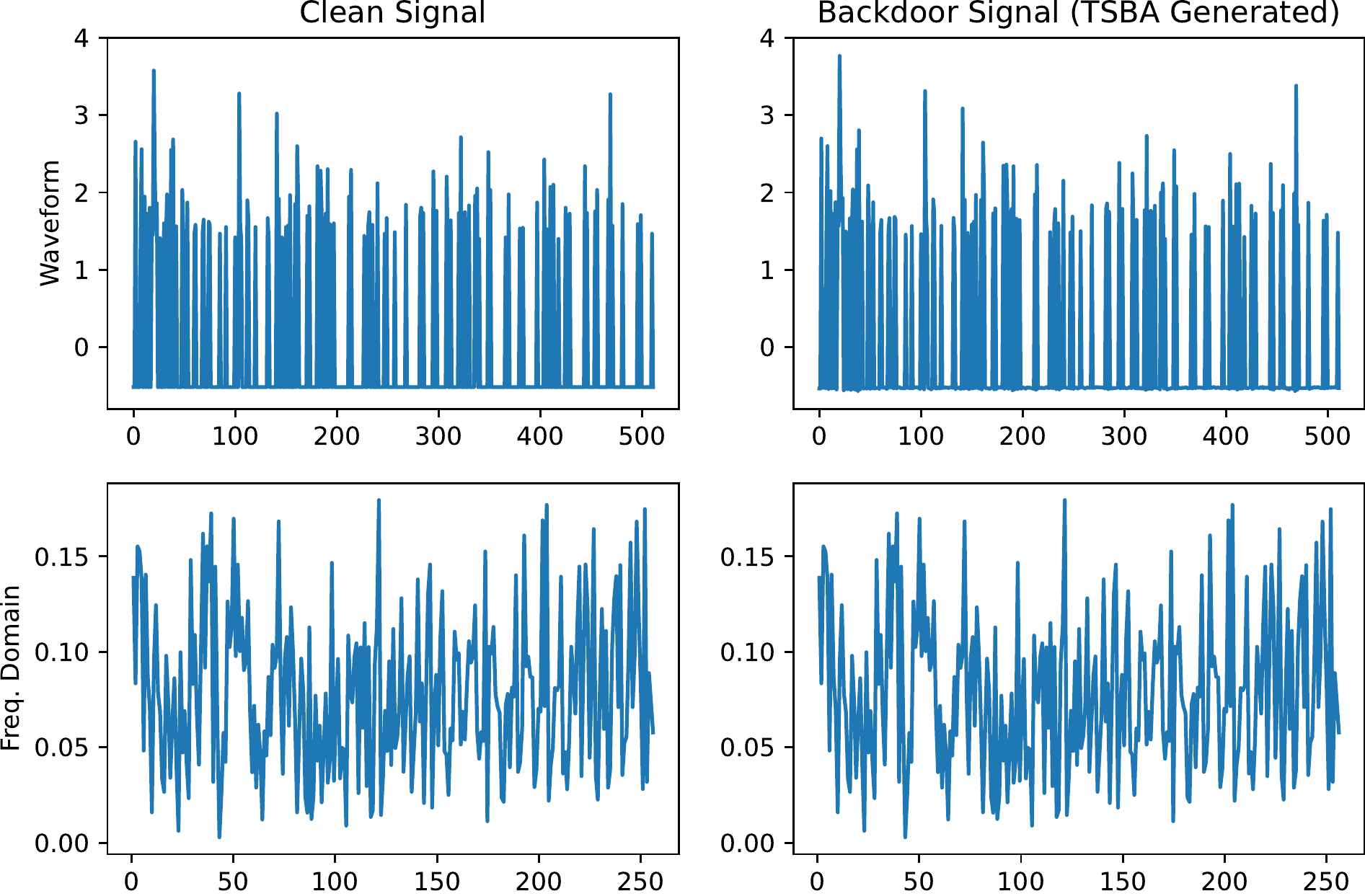}
\end{tabular}
\caption{Example waveform and frequency domain of the original and backdoor time series. The two examples are taken from ($D_{2}$) ECG5000 dataset and ($D_{3}$) Earthquake dataset, respectively.}
\label{fig:example_pattern}
\end{figure*}

We evaluate our TSBA under both threat models \textit{TSBA-A} and \textit{TSBA-B} (introduced in Section \ref{sec:threat_model}) on 13 time series datasets, where 8 of them are univariate datasets from the UCR Archive, and the rest 5 are multivariate datasets from the MTS Archive. Statistics of the 13 datasets can be found in Appendix \ref{appendix:dataset}. Due to the small number of instances and the non-typical train-test split for most datasets, we apply 10-fold cross-validation for all of our experiments and report the average clean accuracy (CA) and attack success rate (ASR).

The poisoning rate for all attacks is set to be 10\%. For better stealthiness, the backdoor trigger pattern is clipped to be 10\% of the signal amplitude ($\vx_{max} - \vx_{min}$) for each sample. For \textit{TSBA-A} and the three baseline methods, we directly take the backdoored classifier as the final classifier. For \textit{TSBA-B}, the final classifier is trained with the 10\%-poisoned training dataset by the trained trigger pattern generator. We select the first class of each dataset as the backdoor class. Details of all network architectures can be found in Appendix \ref{appendix:dataset}. We first warm start the time series classifier by training it for 20 epochs with purely clean samples. Then, we train each model for 500 epochs and apply early stopping to avoid overfitting. The training is terminated when the backdoored classifier achieves an optimal ASR and CA on the validation set according to the 10-fold cross validation.

\subsection{Main Results}

The attack performance of different attacking methods is reported in Table \ref{table:exp_backdoor}.
It is clear that our proposed TSBA-A and TSBA-B attacks achieved the best attack performance among all the baseline attacks, in terms of both ASR and CA. 
Specifically, \textit{TSBA-A} achieves 100\% ASR in 19 out of the 39 experiments, while an average ASR of 98.2\% in the rest of the 20 experiments. The average clean accuracy drop is around 2.45\% and 2.25\% for univariate and multivariate datasets, respectively. 
For \textit{TSBA-B}, it is less powerful than  \textit{TSBA-A} but can still achieve an average ASR of 81.6\%. 
It causes an average clean accuracy drop of 5.09\% and 4.63\% for univariate and multivariate datasets, respectively. 
Surprisingly, as a weaker attack, the CA performance of \textit{TSBA-B} is no better than \textit{TSBA-A}.
We suspect this is because \textit{TSBA-B} has zero control over the training procedure and does not have the benign counterparts of the poisoned samples in the training set. 
Compared with static noise, our \textit{TSBA-B} achieves significantly higher ASR in all experiments while maintaining similar clean accuracy in most cases. 
These results confirm the attack effectiveness of our proposed TSBA attacks.

\subsection{Time Series vs. Images}\label{sec:understanding}

As shown in Table \ref{table:exp_backdoor}, directly applying image-equivalent backdoor attacks (\textit{i.e.,} the two variants of vanilla backdoor) on time series data fails to deliver strong attack performance as in the image domain. 
This is because time series are generally of lower input dimensions and fewer degrees of freedom  than images, making fixed trigger patterns less effective. 
The fast and steep shifts induced by the vanilla backdoor patterns may not be sensitive to the classifier and could be suppressed by the clean peak or trough signals.
Even for the variant that randomly applies the trigger pattern to cover one or more peaks (or troughs), it could only improve the ASR by an average of 1.63\% while reducing the clean accuracy by an average of 2.53\%. 
In fact, effective trigger patterns should produce smooth transitions across the entire signal. By simply adding the static powerline noise with a small 10\% (standardized) amplitude, it achieves  significantly higher ASRs and CAs compared with the vanilla backdoor. 
Our TSBA attacks present a more advanced version of the static noise that can generate more smooth and persisting trigger patterns, and more importantly, be adaptive to each input time series.

\subsection{Stealthiness Analysis}

We use the root mean square (RMS) of the trigger patterns to measure their stealthiness. We apply the trained trigger generator to create sample-specific trigger patterns for all samples in each of the 13 datasets, then report their mean RMS values within each dataset as well as the average overall 13 datasets. To simplify the results, this experiment was conducted under the TSBA-A threat model based on the FCN, one most commonly used model architectures for time series classification. 
The results are reported in Table \ref{table:stealth}.

\begin{table}[h!]
\setlength{\tabcolsep}{0.7em}
\centering
\caption{The RMS of the generated trigger patterns by TSBA. All RMS values are normalized with respect to the magnitude of each sample. $\text{RMS}_{\text{Top 1\%}}$ only computes the highest 1\% portion sorted by the absolute signal values.}
\begin{tabular}{lccccccc}
\toprule
 & $D_{1}$ & $D_{2}$ & $D_{3}$ & $D_{4}$ & $D_{5}$ & $D_{6}$ & $D_{7}$ \\ \midrule
$\text{RMS}_{\text{All}}$ &  0.014 & 0.011 & 0.008 & 0.017 & 0.021 & 0.016 & 0.012 \\
$\text{RMS}_{\text{Top 1\%}}$ & 0.056 & 0.044 & 0.061 & 0.072 & 0.049 & 0.064 & 0.039 \\ \midrule \midrule

 &   $D_{8}$ & $D_{9}$ & $D_{10}$ & $D_{11}$ & $D_{12}$ & $D_{13}$  & \textbf{Avg.} \\ \midrule
$\text{RMS}_{\text{All}}$  &  0.029 & 0.031 & 0.017 & 0.032 & 0.024 & 0.036  &  \textbf{0.021}  \\
$\text{RMS}_{\text{Top 1\%}}$  &  0.052 & 0.087 & 0.068 & 0.047 & 0.059 & 0.084  &  \textbf{0.060}  \\ \bottomrule
\end{tabular}
\label{table:stealth}
\end{table}

\begin{figure*}[ht]
\setlength{\tabcolsep}{0.4em}
\centering
\begin{tabular}{llll}
\includegraphics[width=0.48\columnwidth]{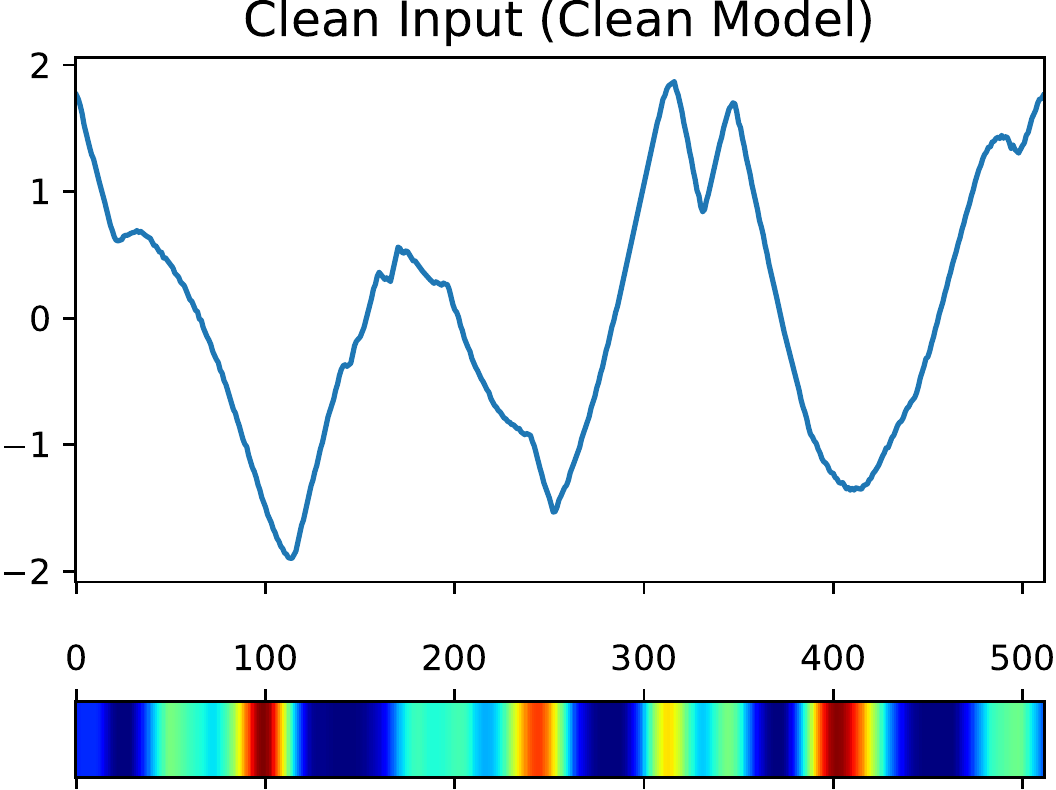} &
\includegraphics[width=0.48\columnwidth]{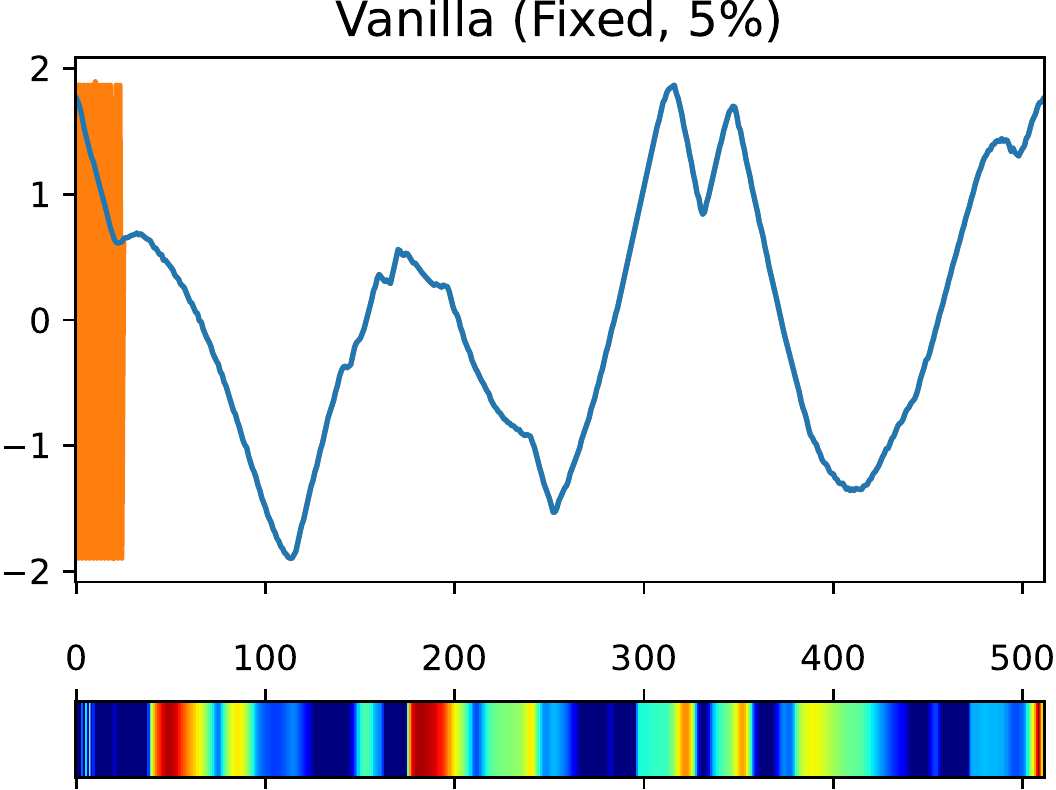} &
\includegraphics[width=0.48\columnwidth]{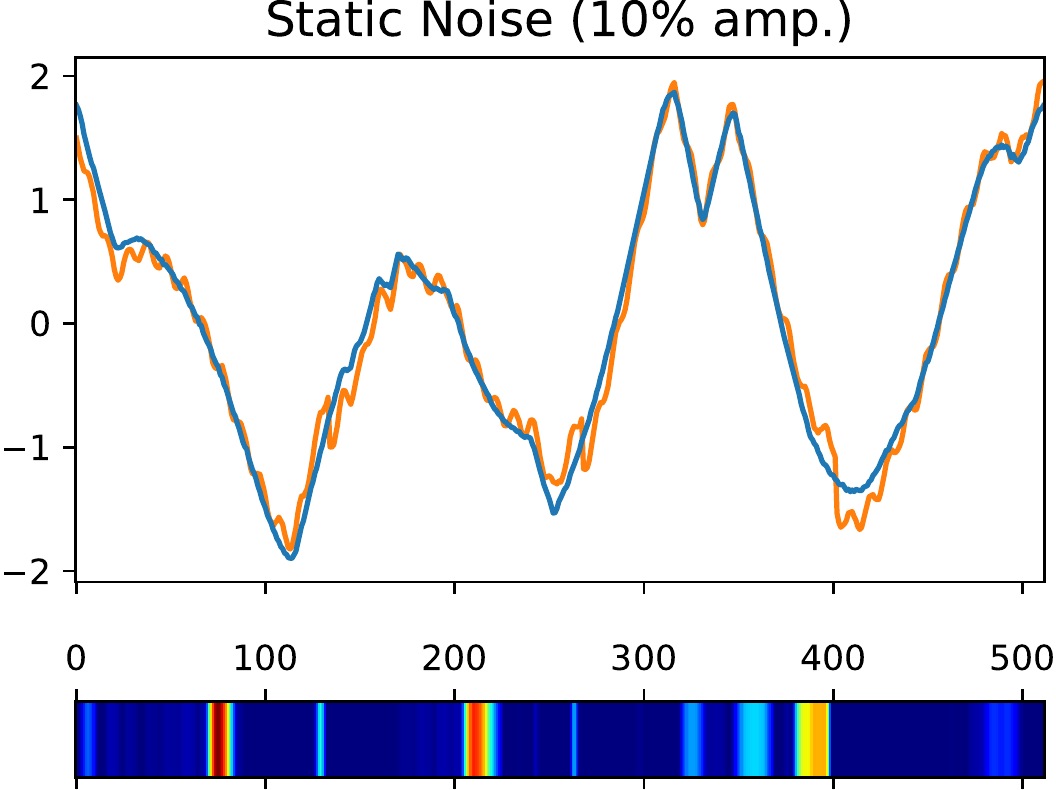} &
\includegraphics[width=0.48\columnwidth]{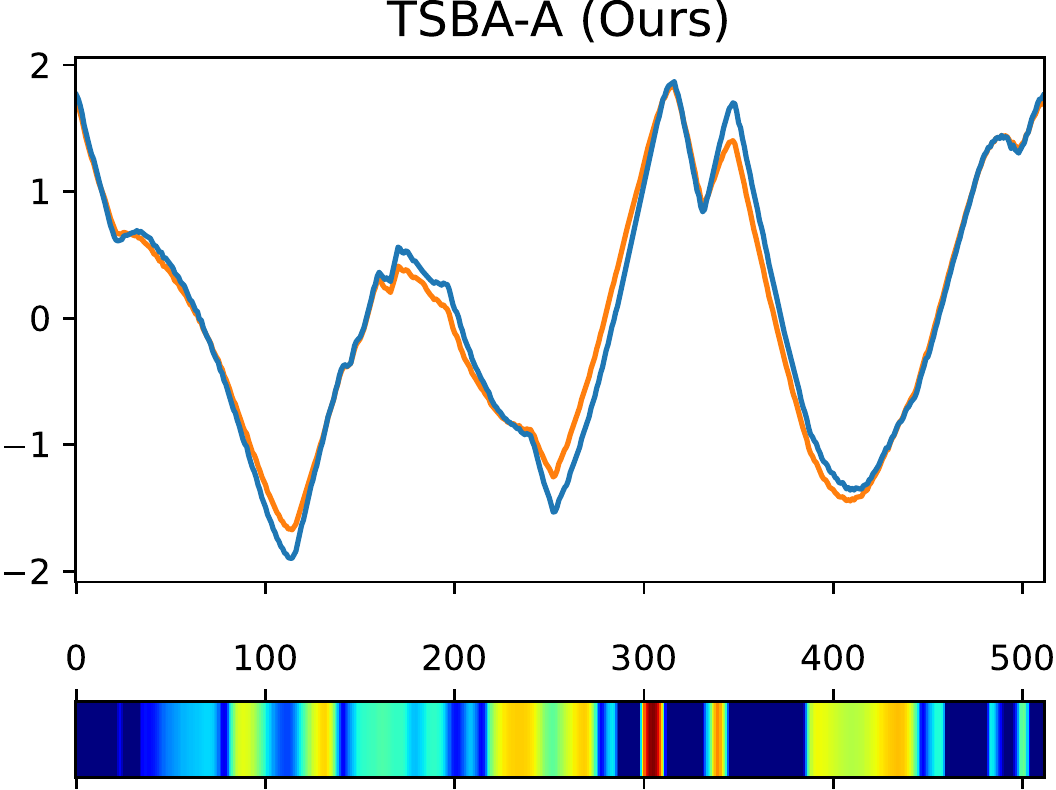} \vspace{1.5ex} \\ 
\includegraphics[width=0.48\columnwidth]{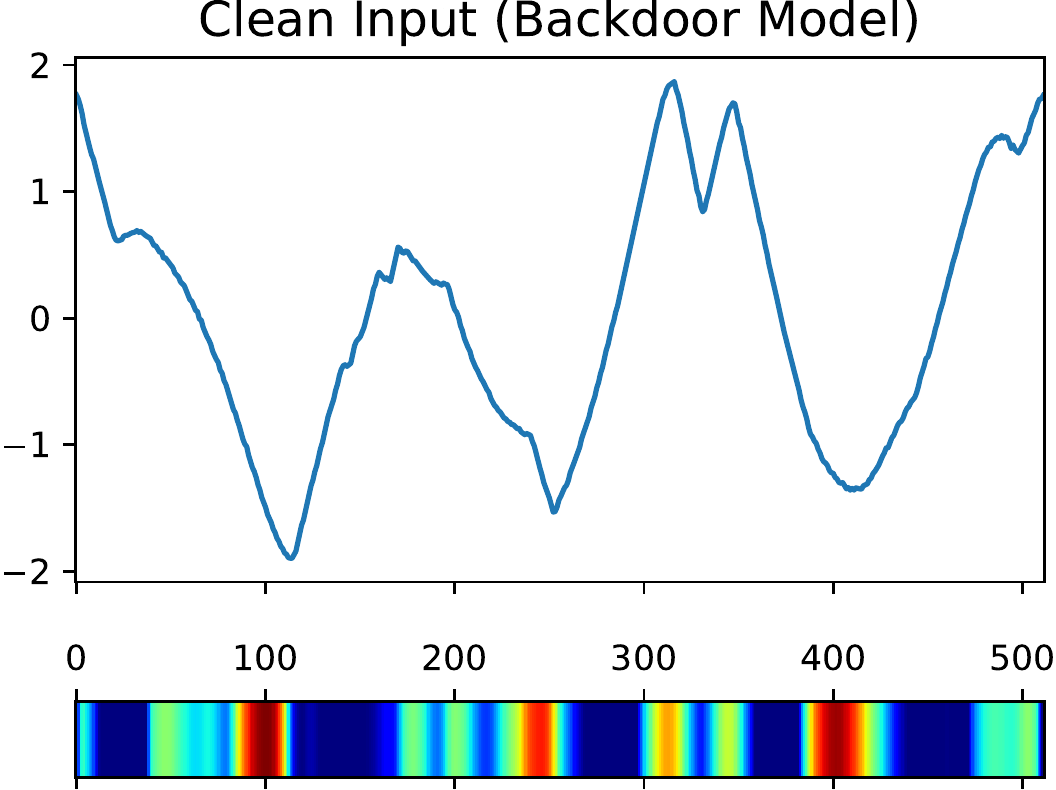} &
\includegraphics[width=0.48\columnwidth]{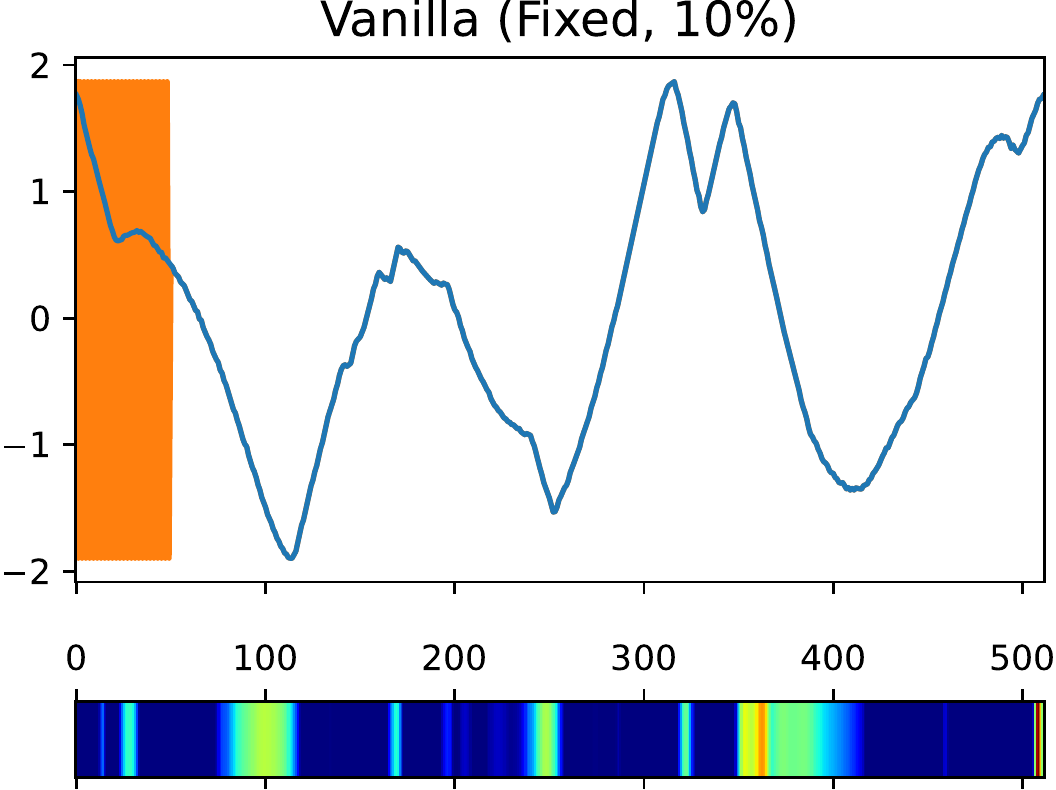} & 
\includegraphics[width=0.48\columnwidth]{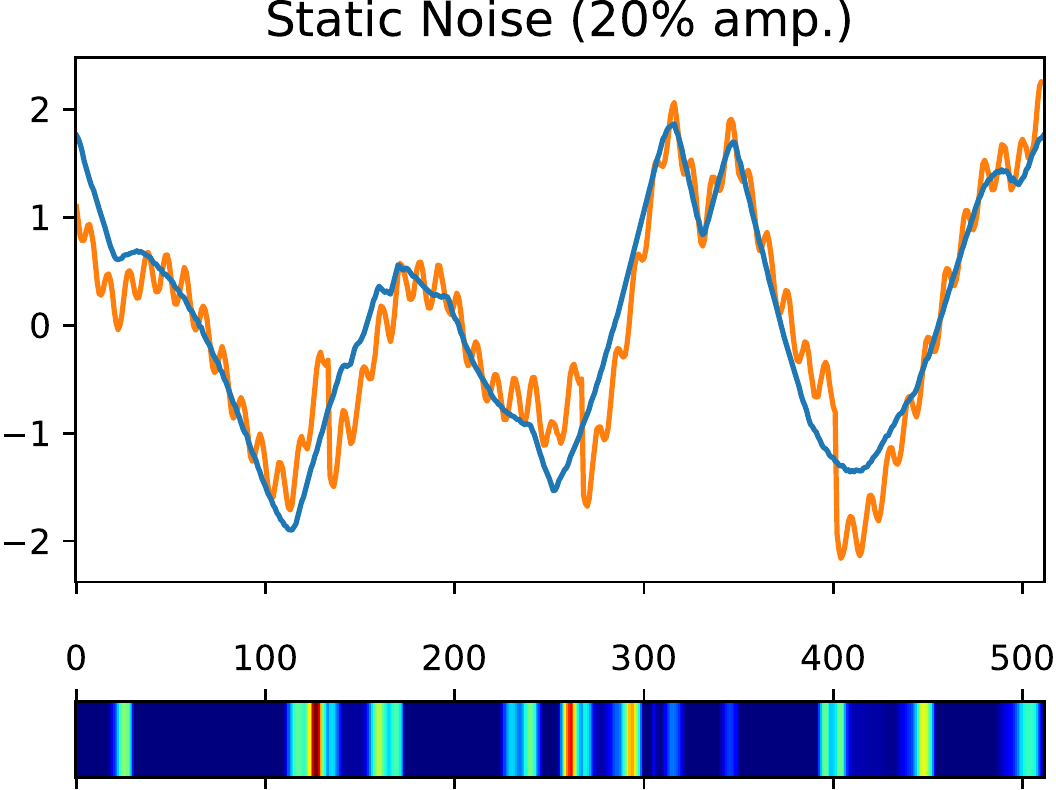} &
\includegraphics[width=0.48\columnwidth]{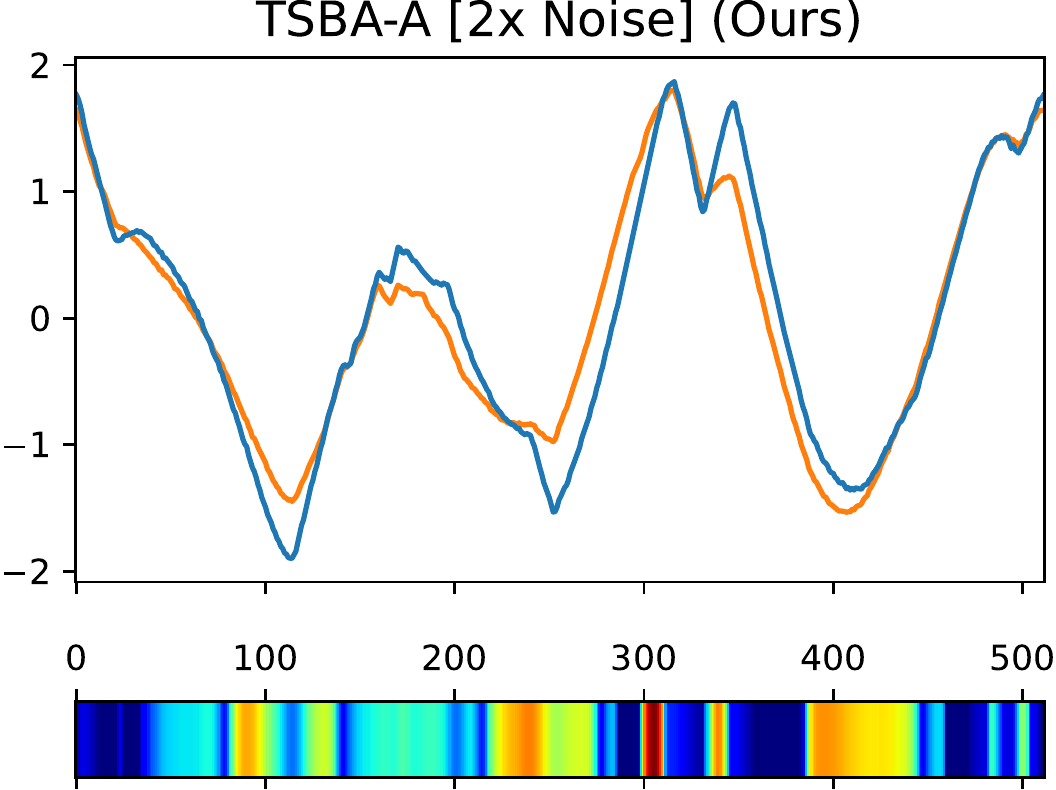}
\end{tabular}
\caption{Waveforms and Grad-CAM attention maps under different attack settings. The example is taken from ($D_{1}$) BirdChicken dataset. The original (blue) and backdoor (orange) signals are plotted together as a comparison.}
\label{fig:grad-cam}
\end{figure*}

It is evident that our TSBA attack introduces very small variations to the original signal, only incurring an average of 2.1\% of the amplitude for all samples across the 13 datasets, although we allow up to 10\% perturbation of each sample's amplitude.
It is worth mentioning that the static noise incurs an average of 10\% of the amplitude, which is much higher than TSBA. The low top 1\% RMS values indicate that only a small portion of each sample is heavily perturbed, while the rest are insignificantly modified. However, such small but sample-specific perturbations can achieve remarkably high ASR with similar clean accuracy.

Figure \ref{fig:example_pattern} shows that the backdoor samples and their benign counterparts are imperceptibly different from their clean counterparts in both the time domain (\textit{i.e.,} waveform) and frequency domain (obtained via Fourier transformation). Figure \ref{fig:grad-cam} shows the different trigger patterns crafted by different attacks. The trigger patterns generated by our attack are very smooth and visually more natural compared to other baselines. Note that the vanilla backdoor attack creates a high variation block at the beginning of the signal, while the static noise causes a suspicious sawtooth effect across the entire signal.
This confirms the high stealthiness of TSBA and the benefit of our generative approach. Our proposed TSBA attacks are capable of generating stealthy backdoor triggers owing to their generative design. 

\subsection{Grad-CAM Visualization}

To help understand the working mechanism of the trigger patterns generated by TSBA, we visualize the Grad-CAM \cite{selvaraju2017grad} attention map in Figure \ref{fig:grad-cam} for the trigger patterns crafted by different attacks and their boosted versions. The boosted triggers can help visualize the differences. 
For the vanilla backdoor with a fixed pattern position (at the beginning), we double its perturbation magnitude from 5\% to 10\%. For the static noise, we directly double the amplitude of the added noise pattern. For our \textit{TSBA-A}, we retrain the model with a doubled clipping rate $\xi$, from 10\% to 20\%. Amongst all six backdoored samples (with their corresponding models), only our TSBA-A and its boosted version have successfully performed the backdoor attack. Unsurprisingly, the trigger for the vanilla backdoor does not activate the hidden backdoor as its attention map is very similar to that of the original signal. For the static noise, there is no obvious region of the altered attention map that can cause incorrect prediction. For the enhanced version of the above baseline attacks, more less-focused regions are shown in the attention map (\textit{i.e.,} more blue regions and less yellow or red regions), indicating that those trigger sequences divert the attention of the classifier. Yet, they are not strong enough to mislead the model's prediction. By contrast, with small-amplitude triggers, our TSBA can produce more disputing regions in the classifier's attention map.

\begin{table*}[h!]
\setlength{\tabcolsep}{1.0em}
\renewcommand{\arraystretch}{1.1}
\centering
\caption{Performance deviation of the baselines and our TSBA attacks against three backdoor defenses. All models use FCN as the time series classifier. The $\Delta$CA and $\Delta$ASR are calculated based on the results from Table \ref{table:exp_backdoor}.}
\begin{tabular}{lcrrrrrrrrrr}
\toprule
\multicolumn{1}{l}{}  \makecell{\multirow{2}{*}{\vspace{-1.5ex}\textbf{Dataset}}}  &   \multirow{2}{*}{\vspace{-1.5ex}\textbf{Defense}}   & \multicolumn{2}{c}{Vanilla (Fixed)}      & \multicolumn{2}{c}{Vanilla (Random)}             & \multicolumn{2}{c}{Static Noise}                    & \multicolumn{2}{c}{\textbf{TSBA-A (Ours)}}              & \multicolumn{2}{c}{\textbf{TSBA-B (Ours)}}          \\  \cmidrule{3-12}
 & & \textbf{$\Delta$CA} & \textbf{$\Delta$ASR} & \textbf{$\Delta$CA} & \textbf{$\Delta$ASR} & \textbf{$\Delta$CA} & \textbf{$\Delta$ASR} & \textbf{$\Delta$CA} & \textbf{$\Delta$ASR} & \textbf{$\Delta$CA} & \textbf{$\Delta$ASR} \\ \midrule

\multirow{3}{*}{\begin{tabular}[c]{@{}c@{}} ($D_{1}$) BirdChicken\end{tabular}} 
   &  NC    & +1.0\%    & -52.0\%    & +2.0\%    & -56.0\%    & -1.0\%    & -27.0\%    & 0.0\%    & -1.0\%    & -1.0\%    & -3.0\%  \\
   &  FP    & -1.0\%    & -51.0\%    & -1.0\%    & -47.0\%    & -1.0\%    & -71.0\%    & -2.0\%    & -3.0\%    & -0.5\%    & -6.0\%  \\
   &  ANP    & -0.0\%    & -52.0\%    & -0.5\%    & -56.0\%    & -1.0\%    & -76.0\%    & -1.0\%    & -6.0\%    & -1.0\%    & -7.0\%  \\ \midrule
   
\multirow{3}{*}{\begin{tabular}[c]{@{}c@{}} ($D_{2}$) ECG5000\end{tabular}}
   &  NC    & +1.5\%    & -56.1\%    & +2.1\%    & -55.4\%    & -1.4\%    & -32.2\%    & -0.4\%    & -1.1\%    & -3.7\%    & -5.2\%  \\
   &  FP    & -1.1\%    & -55.2\%    & -0.8\%    & -48.9\%    & -1.0\%    & -66.4\%    & -2.1\%    & -2.9\%    & -1.7\%    & -6.1\%  \\
   &  ANP    & -0.8\%    & -56.1\%    & -0.6\%    & -55.9\%    & -0.7\%    & -70.1\%    & -1.4\%    & -6.8\%    & -1.5\%    & -8.4\%  \\ \midrule
   
\multirow{3}{*}{\begin{tabular}[c]{@{}c@{}} ($D_{3}$) Earthquakes\end{tabular}}
   &  NC    & +0.5\%    & -51.7\%    & -0.1\%    & -53.7\%    & +0.1\%    & -30.2\%    & -0.1\%    & -0.8\%    & -1.7\%    & -5.4\%  \\
   &  FP    & -0.9\%    & -50.4\%    & -1.2\%    & -48.1\%    & -2.1\%    & -61.9\%    & -2.7\%    & -2.6\%    & -2.2\%    & -7.7\%  \\
   &  ANP    & -0.2\%    & -51.7\%    & -0.4\%    & -55.8\%    & -1.2\%    & -71.3\%    & -1.2\%    & -5.9\%    & -1.4\%    & -6.9\%  \\ \midrule

\multirow{3}{*}{\begin{tabular}[c]{@{}c@{}} ($D_{4}$) ElectricDevices\end{tabular}}
   &  NC    & +1.9\%    & -47.1\%    & +1.5\%    & -48.1\%    & -0.9\%    & -29.7\%    & -1.0\%    & -1.9\%    & -2.2\%    & -4.7\%  \\
   &  FP    & -0.1\%    & -47.2\%    & -0.6\%    & -49.1\%    & -1.1\%    & -69.6\%    & -3.4\%    & -1.8\%    & -2.7\%    & -5.6\%  \\
   &  ANP    & -1.4\%    & -47.7\%    & -1.0\%    & -49.7\%    & -2.4\%    & -74.1\%    & -1.9\%    & -6.4\%    & -1.7\%    & -8.7\%  \\ \midrule

\multirow{3}{*}{\begin{tabular}[c]{@{}c@{}} ($D_{5}$) Haptics\end{tabular}}
   &  NC    & +0.1\%    & -44.5\%    & +0.4\%    & -48.1\%    & -0.3\%    & -33.4\%    & -0.6\%    & -1.5\%    & -0.8\%    & -3.4\%  \\
   &  FP    & -1.1\%    & -39.8\%    & -0.8\%    & -42.7\%    & -1.5\%    & -61.2\%    & -1.9\%    & -2.7\%    & -2.1\%    & -4.9\%  \\
   &  ANP    & -1.0\%    & -44.0\%    & -1.2\%    & -49.0\%    & -1.8\%    & -67.1\%    & -0.9\%    & -4.9\%    & -1.7\%    & -6.4\%  \\ \midrule
   
\multirow{3}{*}{\begin{tabular}[c]{@{}c@{}} ($D_{6}$) PowerCons\end{tabular}}
   &  NC    & +1.0\%    & -57.9\%    & +0.3\%    & -62.0\%    & -1.4\%    & -35.9\%    & -0.2\%    & -0.4\%    & -0.7\%    & -1.5\%  \\
   &  FP    & -0.6\%    & -51.6\%    & -0.8\%    & -53.4\%    & -1.7\%    & -67.6\%    & -3.0\%    & -2.4\%    & -2.7\%    & -3.1\%  \\
   &  ANP    & -0.4\%    & -58.3\%    & -0.7\%    & -61.6\%    & -1.0\%    & -73.8\%    & -2.1\%    & -6.4\%    & -2.0\%    & -8.2\%  \\ \midrule
   
\multirow{3}{*}{\begin{tabular}[c]{@{}c@{}} ($D_{7}$) ShapeletSim\end{tabular}}
   &  NC    & +1.8\%    & -54.8\%    & -1.6\%    & -56.9\%    & -3.4\%    & -41.6\%    & +0.3\%    & -0.2\%    & -0.4\%    & -1.7\%  \\
   &  FP    & -1.2\%    & -49.7\%    & -0.5\%    & -52.6\%    & -3.1\%    & -68.9\%    & -2.4\%    & -1.9\%    & -2.6\%    & -2.9\%  \\
   &  ANP    & -0.2\%    & -56.3\%    & -0.3\%    & -57.3\%    & -1.9\%    & -74.5\%    & -1.2\%    & -7.1\%    & -1.8\%    & -6.8\%  \\ \midrule

\multirow{3}{*}{\begin{tabular}[c]{@{}c@{}} ($D_{8}$) Wine\end{tabular}}
   &  NC    & -0.1\%    & -51.3\%    & +0.4\%    & -52.0\%    & -0.9\%    & -39.9\%    & -0.1\%    & -1.1\%    & -1.2\%    & -2.0\%  \\
   &  FP    & -1.0\%    & -46.7\%    & -0.7\%    & -46.9\%    & -1.2\%    & -67.5\%    & -1.2\%    & -2.6\%    & -2.4\%    & -4.9\%  \\
   &  ANP    & -0.6\%    & -52.6\%    & -0.4\%    & -53.7\%    & -0.9\%    & -71.6\%    & -1.0\%    & -5.0\%    & -2.1\%    & -7.1\%  \\ \midrule \midrule

\multirow{3}{*}{\begin{tabular}[c]{@{}c@{}} ($D_{9}$) ArabicDigits\end{tabular}}
   &  NC    & +1.4\%    & -52.1\%    & +1.9\%    & -55.0\%    & -1.4\%    & -28.4\%    & -0.8\%    & -1.1\%    & -1.6\%    & -1.4\%  \\
   &  FP    & -0.2\%    & -48.2\%    & -0.8\%    & -51.9\%    & -1.0\%    & -60.1\%    & -4.6\%    & -3.5\%    & -3.8\%    & -3.4\%  \\
   &  ANP    & -0.7\%    & -52.4\%    & -0.4\%    & -56.1\%    & -0.9\%    & -65.2\%    & -2.4\%    & -7.8\%    & -3.7\%    & -8.2\%  \\ \midrule
   
\multirow{3}{*}{\begin{tabular}[c]{@{}c@{}} ($D_{10}$) ECG\end{tabular}}
   &  NC    & +2.2\%    & -58.9\%    & +1.5\%    & -56.3\%    & -2.1\%    & -29.8\%    & -0.9\%    & -2.4\%    & -1.1\%    & -3.1\%  \\
   &  FP    & -0.7\%    & -53.6\%    & -0.8\%    & -52.1\%    & -1.7\%    & -64.2\%    & -3.8\%    & -4.1\%    & -3.9\%    & -4.7\%  \\
   &  ANP    & -0.1\%    & -58.4\%    & -0.2\%    & -58.1\%    & -0.9\%    & -76.9\%    & -2.0\%    & -9.0\%    & -3.2\%    & -8.1\%  \\ \midrule
   
\multirow{3}{*}{\begin{tabular}[c]{@{}c@{}} ($D_{11}$) KickvsPunch\end{tabular}}
   &  NC    & +1.7\%    & -53.1\%    & +2.4\%    & -53.6\%    & -2.4\%    & -39.2\%    & -3.1\%    & -1.6\%    & -3.0\%    & -4.1\%  \\
   &  FP    & -0.9\%    & -51.2\%    & -0.4\%    & -52.0\%    & -2.2\%    & -66.1\%    & -2.5\%    & -2.9\%    & -2.1\%    & -7.9\%  \\
   &  ANP    & -0.5\%    & -55.1\%    & -0.6\%    & -53.9\%    & -1.4\%    & -70.7\%    & -1.6\%    & -8.6\%    & -2.6\%    & -9.1\%  \\ \midrule

\multirow{3}{*}{\begin{tabular}[c]{@{}c@{}} ($D_{12}$) NetFlow\end{tabular}}
   &  NC    & +0.1\%    & -47.6\%    & +0.3\%    & -47.6\%    & -2.6\%    & -25.2\%    & -0.7\%    & -2.9\%    & -1.2\%    & -2.4\%  \\
   &  FP    & -0.2\%    & -38.3\%    & -0.9\%    & -41.8\%    & -2.4\%    & -73.5\%    & -2.9\%    & -4.3\%    & -2.8\%    & -3.8\%  \\
   &  ANP    & -0.4\%    & -48.9\%    & -0.1\%    & -48.8\%    & -1.0\%    & -77.4\%    & -1.5\%    & -8.1\%    & -1.7\%    & -7.6\%  \\ \midrule

\multirow{3}{*}{\begin{tabular}[c]{@{}c@{}} ($D_{13}$) UWave\end{tabular}}
   &  NC    & +0.8\%    & -42.1\%    & +1.9\%    & -45.1\%    & -3.1\%    & -38.7\%    & -1.7\%    & -3.0\%    & -2.1\%    & -2.4\%  \\
   &  FP    & -1.0\%    & -38.9\%    & -1.3\%    & -41.2\%    & -2.0\%    & -58.8\%    & -3.6\%    & -3.7\%    & -4.4\%    & -4.2\%  \\
   &  ANP    & -0.4\%    & -43.7\%    & -0.8\%    & -46.4\%    & -1.4\%    & -62.3\%    & -2.2\%    & -7.7\%    & -3.0\%    & -8.1\%  \\

\bottomrule
\end{tabular}
\vspace{1ex}
\label{table:exp_defense}
\end{table*}

\subsection{Resistance to Backdoor Defenses}
Here, we show that our TSBA attack can easily evade state-of-the-art backdoor defense methods, including Neural Cleanse (NC) \cite{wang2019neural}, Fine-Pruning (FP) \cite{liu2018fine} and Adversarial Neuron Pruning (ANP) \cite{wu2021adversarial}. The results of our TSBA-A/B and the three baselines are reported in Table \ref{table:exp_defense}.

\subsubsection{Evading NC}

NC \cite{wang2019neural} can not only detect whether a DNN has been backdoored but also recover and mitigate the backdoor trigger via reverse engineering and unlearning. Here, we apply NC with unlearning that allows the model to decide which weights are problematic and should be updated through training. This defense can significantly eliminate the two Vanilla baseline attacks, reducing their ASR by $42\%$-$62\%$. However, it fails on the other 3 attacks due to its limitation on detecting static trigger patterns that are not applied to the entire input. Since our TSBAs use sample-specific triggers, NC can only reduce their ASRs by 1\%-6\%.

\subsubsection{Evading FP}

FP \cite{liu2018fine} exploits the advantages of both pruning and fine-tuning, and progressively removes the dormant neurons that are conditioned on clean images to mitigate the backdoor. We apply FP with the default hyperparameters on the backdoored classifiers trained by our \textit{TSBA} attacks. We set the pruning rates to 30\%. Our TSBA demonstrates high resistance to the FP defense with less than 5\% and 9\% ASR drop for TSBA-A and TSBA-B, respectively. This implies that the trigger patterns generated by TSBAs are deeply mixed into the clean signals in the representation space, making it hard to be removed by either neuron pruning or clean-data based fine-tuning.

\subsubsection{Evading ANP}

ANP \cite{wu2021adversarial} is a recent defense method that removes potential backdoors by pruning the most susceptible neurons to adversarial perturbations. ANP demonstrates state-of-the-art defense performance against a number of backdoor attacks. Here, we apply ANP to all backdoored models with a perturbation budget $\epsilon=0.4$, trade-off coefficient $\alpha=0.2$, and constant learning rate 0.2.
All neuron masks are optimized using Stochastic Gradient Descent (SGD). ANP can significantly eliminate the two Vanilla attacks (i.e., Fixed and Random), reducing their ASRs to $<$2\%. For the static noise attack, ANP brings its ASR down to less than 10\% on all datasets. Our TSBA attacks are fairly robust against this strong defense with ASR drops by $<9\%$.

The ASRs of the three baseline attacks are all below 10\% under the backdoor defenses. However, the ASRs of our TSBA attacks drop only by less than 10\% across all 3 defenses and 13 datasets, which are still above 80\%. This proves that, even under strong backdoor defenses, our attacks can still pose high threats.
It is also worth mentioning that, since our triggers are dynamic and sample-specific, existing defense techniques designed for static patterns may not apply to our attacks.

\section{Universal Trigger Generator}
\label{sec:universal_gen}
To evaluate the attack performance of our proposed universal trigger generator, here we train the model on a combined dataset of 10 \emph{new} univariate datasets from UCR Archive, namely the first 11 datasets except for $D_{1}$ BirdChicken (as $D_{1}$ are selected in our experiments). 
The training of the generator is early stopped when an optimal ASR (on the training set) is achieved.
Then, we use the obtained universal trigger generator to create poisoned training set for each of the 13 datasets used in our previous experiments, with a poison rate of 10\%. For multivariate datasets, we apply the universal generator to each time-dependent variable separately. Each backdoored classifier was trained following the same training procedure and setting under our \textit{TSBA-B} threat model (stated in Section \ref{sec:exp_setup}).

\begin{table}[h]
\setlength{\tabcolsep}{0.7em}
\renewcommand{\arraystretch}{1.05}
\centering
\caption{Performance of our universal trigger generator. The differences in CA and ASR are calculated versus TSBA-B.}
\begin{tabular}{ccc|cccc}
\toprule
\textbf{Dataset} &   \textbf{Classifier}   &  \textbf{Clean}    &   \textbf{CA}   &   \textbf{ASR}     &   \textbf{$\Delta$CA }   &   \textbf{$\Delta$ASR}    \\ \midrule

\multirow{3}{*}{\begin{tabular}[c]{@{}c@{}} $D_{1}$\end{tabular}}  &  
   TCN       & 96.0\%    & 94.0\%    & 92.0\%    & +2.0\%    & -2.0\%    \\
& ResNet  & 92.0\%    & 90.0\%    & 92.0\%     & +4.0\%    & -2.0\%   \\
& LSTM     & 94.0\%    & 89.0\%    & 88.0\%     & +1.0\%    & -4.0\%   \\ \midrule

\multirow{3}{*}{\begin{tabular}[c]{@{}c@{}} $D_{2}$\end{tabular}}  &  
   TCN       & 94.6\%    & 90.1\%    & 87.5\%    & +0.6\%    & -1.5\%    \\
& ResNet  & 93.9\%    & 90.2\%    & 85.5\%     & +1.2\%    & -0.7\%   \\
& LSTM     & 94.1\%    & 88.8\%    & 82.2\%    & -0.3\%    & -1.9\%    \\ \midrule

\multirow{3}{*}{\begin{tabular}[c]{@{}c@{}} $D_{3}$\end{tabular}}  &  
   TCN       & 72.5\%   & 69.9\%    & 78.6\%    & +3.0\%    & -2.5\%    \\
& ResNet  & 71.7\%    & 68.5\%    & 75.7\%    & +2.2\%    & -2.7\%    \\
& LSTM     & 72.2\%   & 67.7\%    & 73.2\%    & +1.7\%    & -3.6\%    \\ \midrule

\multirow{3}{*}{\begin{tabular}[c]{@{}c@{}} $D_{4}$\end{tabular}}  &  
   TCN       & 72.3\%    & 69.2\%    & 80.6\%    & +2.1\%    & -1.9\%    \\
& ResNet  & 73.4\%    & 71.0\%    & 78.5\%    & +2.4\%    & -3.1\%    \\
& LSTM     & 70.8\%    & 68.9\%    & 79.5\%    & +3.0\%    & -2.4\%    \\ \midrule

\multirow{3}{*}{\begin{tabular}[c]{@{}c@{}} $D_{5}$\end{tabular}}  &  
   TCN       & 50.2\%    & 50.6\%    & 86.1\%    & +4.1\%    & -0.6\%    \\
& ResNet  & 51.6\%    & 50.6\%    & 87.1\%    & +2.2\%    & -1.1\%    \\
& LSTM     & 54.0\%    & 51.0\%    & 84.9\%    & +1.4\%    & -3.7\%    \\ \midrule

\multirow{3}{*}{\begin{tabular}[c]{@{}c@{}} $D_{6}$\end{tabular}}  &  
   TCN       & 88.2\%    & 82.3\%    & 79.3\%    & +0.6\%    & +0.2\%    \\
& ResNet  & 89.7\%    & 84.1\%    & 76.4\%     & -0.2\%    & -1.8\%    \\
& LSTM     & 86.4\%     & 80.4\%    & 75.0\%    & +1.1\%    & -0.6\%    \\ \midrule

\multirow{3}{*}{\begin{tabular}[c]{@{}c@{}} $D_{7}$\end{tabular}}  &  
   TCN       & 72.4\%    & 67.8\%    & 83.6\%    & +1.3\%    & -1.1\%    \\
& ResNet  & 77.9\%     & 72.3\%    & 83.5\%    & +0.8\%    & -1.7\%    \\
& LSTM     & 68.5\%    & 63.0\%    & 81.0\%    & +2.6\%    & -3.4\%    \\ \midrule

\multirow{3}{*}{\begin{tabular}[c]{@{}c@{}} $D_{8}$\end{tabular}}  &  
   TCN       & 59.6\%    & 61.1\%    & 81.8\%    & +4.6\%    & +1.1\%    \\
& ResNet  & 74.5\%    & 73.2\%    & 75.3\%    & +1.3\%    & -1.6\%    \\
& LSTM     & 66.8\%    & 67.2\%    & 75.9\%    & +3.1\%    & -2.3\%    \\ \midrule \midrule

\multirow{3}{*}{\begin{tabular}[c]{@{}c@{}} $D_{9}$\end{tabular}}  &  
   TCN       & 99.4\%    & 96.9\%    & 75.8\%    & +2.7\%    & -1.6\%    \\
& ResNet  & 99.6\%    & 95.7\%    & 76.7\%    & +1.5\%    & -2.2\%    \\
& LSTM     & 94.2\%    & 90.6\%    & 72.1\%    & +2.4\%    & -2.9\%    \\ \midrule

\multirow{3}{*}{\begin{tabular}[c]{@{}c@{}} $D_{10}$\end{tabular}}  &  
   TCN       & 87.4\%    & 83.3\%    & 79.9\%    & +0.5\%    & -1.7\%    \\
& ResNet  & 86.2\%    & 80.6\%    & 83.5\%    & -0.9\%    & -0.2\%    \\
& LSTM     & 86.8\%    & 84.8\%    & 77.0\%    & +1.8\%    & -3.5\%    \\ \midrule

\multirow{3}{*}{\begin{tabular}[c]{@{}c@{}} $D_{11}$\end{tabular}}  &  
   TCN       & 54.0\%    & 52.5\%    & 79.7\%    & +1.5\%    & +2.6\%    \\
& ResNet  & 51.3\%    & 52.5\%    & 76.7\%    & +3.8\%    & +0.9\%    \\
& LSTM     & 51.1\%    & 51.2\%    & 76.5\%    & +2.4\%    & +1.1\%    \\ \midrule

\multirow{3}{*}{\begin{tabular}[c]{@{}c@{}} $D_{12}$\end{tabular}}  &  
   TCN       & 89.4\%    & 86.0\%    & 82.6\%    & +1.5\%    & +0.2\%    \\
& ResNet  & 77.5\%    & 76.0\%    & 83.9\%    & +3.1\%    & -0.6\%    \\
& LSTM     & 88.6\%    & 87.1\%    & 83.5\%    & +3.7\%    & -1.4\%    \\ \midrule

\multirow{3}{*}{\begin{tabular}[c]{@{}c@{}} $D_{13}$\end{tabular}}  &  
   TCN       & 93.4\%    & 89.9\%    & 69.4\%    & +2.6\%    & -3.1\%    \\
& ResNet  & 92.2\%    & 87.7\%    & 75.1\%    & +1.2\%    & -1.6\%    \\
& LSTM     & 84.1\%    & 80.7\%    & 68.6\%    & +2.0\%    & -0.8\%    \\

\bottomrule
\end{tabular}
\label{table:exp_universal}
\end{table}

As shown in Table \ref{table:exp_universal}, the universal trigger generator exhibits strong performance on all 13 datasets. The clean accuracy is generally higher than our non-universal \textit{TSBA-B} attack, while the attack success rate drops by less than 4\%. 
Note that this attack performance still outperforms the static noise, the strongest baseline attack.
The sample-wise trigger patterns generated by the universal generator are still dynamic and equally stealthy as those generated by \textit{TSBA-A} and \textit{TSBA-B}. Moreover, compared with universal adversarial noises that can be effectively detected and erased by several defense techniques, our universal trigger patterns on time series data are robust to existing image-based backdoor defenses. Our universal trigger generator provides a cheap but extremely effective approach for attacking any type of time series, posing severe threats to deep learning based time series classification.

\section{Conclusion}
In this paper, we proposed a novel generative approach called Time Series Backdoor Attack (TSBA)  for backdoor attacks on time series. TSBA is capable of generating highly stealthy and sample-specific trigger patterns for time series backdoor attacks under two different threat models. We studied the difference between image backdoors vs. time series backdoors via three proposed baseline attacks with fixed patterns. We found that the low dimension and fewer degrees of freedom (due to time dependence) nature of time series data make fixed patterns hardly work as backdoor trigger patterns. We empirically show on 13 representative datasets that our proposed TSBA attacks can achieve high ASRs with small drops in clean accuracy, outperforming all three proposed baseline attacks. 
We analyzed the stealthiness of our attacks as well as their resistance to state-of-the-art backdoor defense methods. We also demonstrate the strong robustness of TSBA to several advanced backdoor defenses. Moreover, our proposed universal trigger generator was also demonstrated to be as effective as dataset-specific TSBA attacks, posing serious threats to deep learning based time series models. 
Although there are still many unknowns to explore in time series backdoor attacks, our work provides a strong baseline and a good starting point for future research in this area. For future work, we aim to expand our current research and design advanced backdoor defense techniques for time series.

\bibliographystyle{ieeetr}
\bibliography{ref}

\appendices

\section{Details of the datasets and network architectures}
\label{appendix:dataset}

\begin{table}[H]
\setlength{\tabcolsep}{0.4em}
\centering
\caption{Statistics of the selected univariate datasets from the UCR Archive.}
\begin{tabular}{l@{\hspace{8pt}}lccccc}
\toprule
\textbf{Dataset} & \textbf{\begin{tabular}[c]{@{}c@{}}Data\\ Type\end{tabular}}  & \textbf{\#Class} & \textbf{\begin{tabular}[c]{@{}c@{}}Frame\\ Length\end{tabular}} & \textbf{\begin{tabular}[c]{@{}c@{}}Training\\ Samples\end{tabular}} & \textbf{\begin{tabular}[c]{@{}c@{}}Test\\ Samples\end{tabular}} \\ \midrule
($D_{1}$) BirdChicken & \textit{Image} & 2 & 512 & 20 & 20 \\
($D_{2}$) ECG5000 & \textit{ECG} & 5 & 140 & 500 & 4500 \\
($D_{3}$) Earthquakes & \textit{Sensor} & 2 & 512 & 322 & 139 \\
($D_{4}$) ElectricDevices & \textit{Device} & 7 & 96 & 8926 & 7711 \\
($D_{5}$) Haptics & \textit{Motion} & 5 & 1092 & 155 & 308 \\ 
($D_{6}$) PowerCons & \textit{Power}  & 2 & 144 & 180 & 180 \\
($D_{7}$) ShapeletSim & \textit{Simulated} & 2 & 500 & 20 & 180 \\
($D_{8}$) Wine & \textit{Spectro} & 2 & 234 & 57 & 54 \\
\bottomrule
\end{tabular}
\label{table:dataset_ucr}
\end{table}

\begin{table}[H]
\setlength{\tabcolsep}{0.5em}
\centering
\caption{Statistics of the selected multivariate datasets from the MTS Archive.}
\begin{tabular}{l@{\hspace{8pt}}cccccc}
\toprule
\textbf{Dataset} & \textbf{\#Class} & \textbf{\#Var.} & \textbf{\begin{tabular}[c]{@{}c@{}}Frame\\ Length\end{tabular}} & \textbf{\begin{tabular}[c]{@{}c@{}}Training\\ Samples\end{tabular}} & \textbf{\begin{tabular}[c]{@{}c@{}}Test\\ Samples\end{tabular}} \\ \midrule
($D_{9}$) ArabicDigits & 10 & 13 & 4--93 & 6600 & 2200 \\
($D_{10}$) ECG & 2 & 2 & 39--152 & 100 & 100 \\
($D_{11}$) KickvsPunch & 2 & 62 & 274--841 & 16 & 10 \\
($D_{12}$) NetFlow & 2 & 4 & 50--997 & 803 & 534 \\
($D_{13}$) UWave & 8 & 3 & 315 & 200 & 4278 \\
\bottomrule
\end{tabular}
\label{table:dataset_mts}
\end{table}

\begin{table}[H]
\centering
\caption{Architectures of the trigger generator network, where ``L" denotes the length of the time series input, and ``D" denotes the number of time-dependent variables of the time series input.}
\begin{tabular}{cccc}
\toprule
\textbf{Layers} (Activation) & \textbf{Kernel size} & \textbf{\# Kernels} & \textbf{Output size} \\ \midrule
\textbf{Input} & - & - & (L, D) \\
\textbf{Conv1D} (ReLU) & 15*1 & 128*D & (L, 128*D) \\
\textbf{Conv1D} (ReLU) & 21*1 & 512*D & (L, 512*D) \\
\textbf{FC} (ReLU) & - & 256 & (L, 256*D) \\
\textbf{FC} (tanh) & - & D & (L, D) \\
\bottomrule
\end{tabular}
\label{table:noise_gen}
\end{table}

\begin{table}[H]
\centering
\caption{Architecture of the universal noise generator.}
\begin{tabular}{cccc}
\toprule
\textbf{Layers} (Activation) & \textbf{Kernel size} & \textbf{\# Kernels} & \textbf{Output size} \\ \midrule
\textbf{Input} & - & - & (L, D) \\
\textbf{Conv1D} (ReLU) & 15*1 & 128*D & (L, 128*D) \\
\textbf{Conv1D} (ReLU) & 21*1 & 512*D & (L, 512*D) \\
\textbf{Conv1D} (ReLU) & 8*1 & 1024*D & (L, 1024*D) \\
\textbf{FC} (ReLU) & - & 512 & (L, 512*D) \\
\textbf{FC} (tanh) & - & D & (L, D) \\
\bottomrule
\end{tabular}
\label{table:uni_gen}
\end{table}

\end{document}